\documentclass[acmtog]{acmart}
\usepackage{booktabs} 
\usepackage{enumitem}
\usepackage{fancyhdr}

\usepackage[ruled]{algorithm2e} 

\SetAlFnt{\small}
\SetAlCapFnt{\small}
\SetAlCapNameFnt{\small}
\SetAlCapHSkip{0pt}

\usepackage{multirow}
\usepackage{subcaption}
\usepackage{graphicx}
\usepackage{mathtools}

\renewcommand\footnotetextcopyrightpermission[1]{}
\graphicspath{{./figs/}}

\newcommand{\red}[1]{\textcolor{black}{#1}}

\usepackage[normalem]{ulem}

\DeclareRobustCommand{\thinskip}{\hskip 0.1em\relax}
\def\emdash{---}
\def\d@sh#1#2{\unskip#1\thinskip#2\thinskip\ignorespaces}
\def\Dash{\d@sh\nobreak\emdash}
\def\Ldash{\d@sh\empty{\hbox{\emdash}\nobreak}}
\def\Rdash{\d@sh\nobreak\emdash}

\definecolor{good}{rgb}{1,1,0.6}
\definecolor{better}{rgb}{1,0.8,0.6}
\definecolor{best}{rgb}{1,0.6,0.6}

\begin{document}
\title{3D GAN Inversion for Controllable Portrait Image Animation}

\author{Connor Z. Lin}
\authornote{Both authors contributed equally.}
\author{David B. Lindell}
\authornotemark[1]
\author{Eric R. Chan}
\author{Gordon Wetzstein}
\affiliation{%
  \institution{Stanford University}
  \city{Stanford}
  \state{CA}
  \country{USA}
}
\email{connorzl@stanford.edu}
\email{lindell@stanford.edu}
\thanks{Project webpage: \url{https://www.computationalimaging.org/publications/3dganinversion}}

\renewcommand\shortauthors{Lin, C. Z. et al.}

\begin{abstract}
Millions of images of human faces are captured every single day; but these photographs portray the likeness of an individual with a fixed pose, expression, and appearance. 
Portrait image animation enables the post-capture adjustment of these attributes from a single image while maintaining a photorealistic reconstruction of the subject's \red{likeness or identity}.
Still, current methods for portrait image animation are typically based on 2D warping operations or manipulations of a 2D generative adversarial network (GAN) and lack explicit mechanisms to enforce multi-view consistency.
Thus these methods may significantly alter the \red{identity} of the subject, especially when the viewpoint relative to the camera is changed. 
In this work, we leverage newly developed 3D GANs, which allow explicit control over the pose of the image subject with multi-view consistency.
We propose a supervision strategy to flexibly manipulate expressions with 3D morphable models, and we show that the proposed method also supports editing appearance attributes, such as age or hairstyle, by interpolating within the latent space of the GAN.  
The proposed technique for portrait image animation outperforms previous methods in terms of image quality, \red{identity preservation}, and pose transfer while also supporting attribute editing.
\end{abstract}



\begin{teaserfigure}
\centering
\includegraphics[width=\columnwidth]{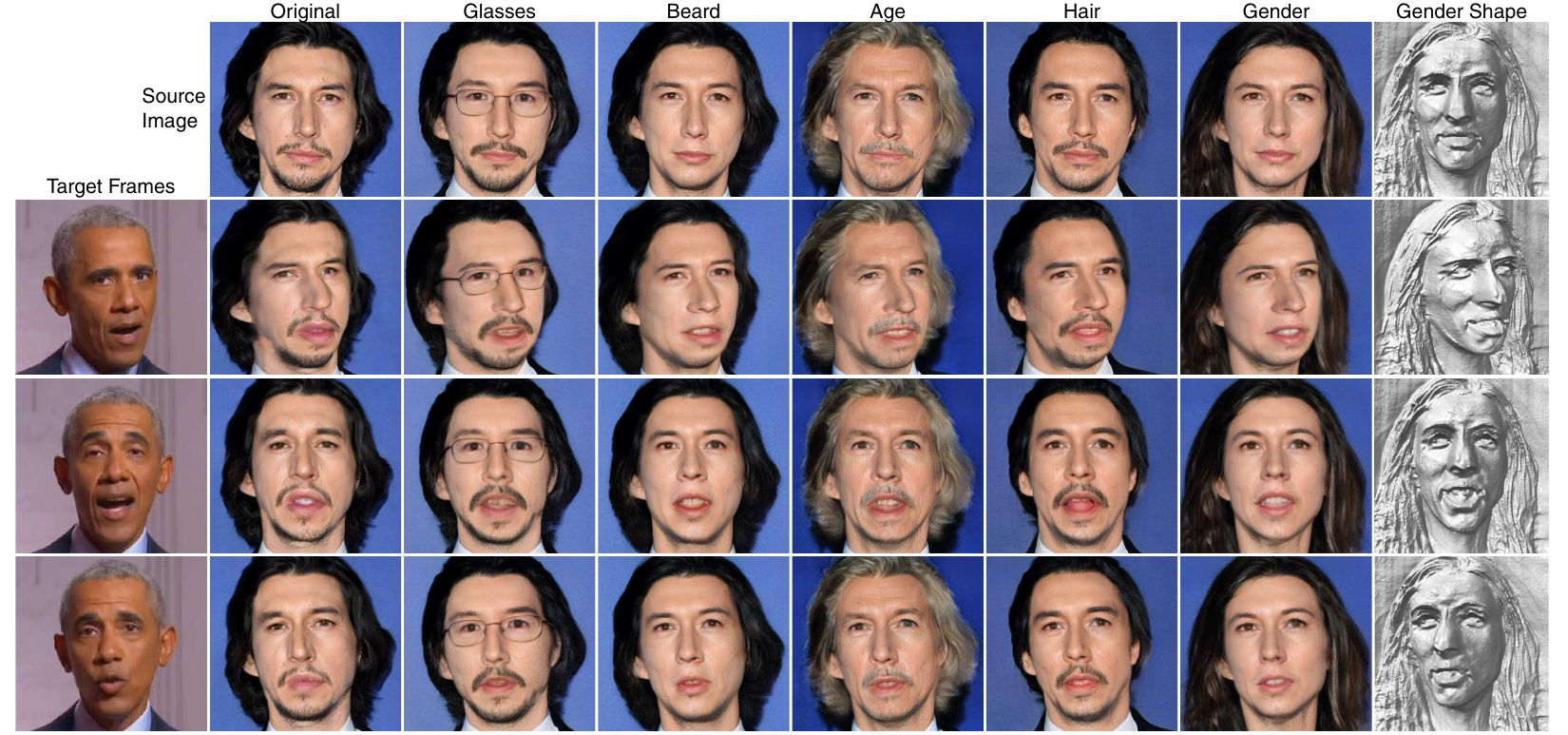}
\caption{Portrait image animation and attribute editing using the proposed technique. Given a source portrait image and a target expression (e.g., specified with a target image), our method transfers the expression and pose to the input source image. We achieve multi-view consistent edits of pose by embedding the expression-edited portrait image into the latent space of a 3D GAN (see predicted underlying shape, right). By interpolating within the latent space of the GAN, we can also apply our method to animate attribute-edited images, allowing adjustments to age, hair, gender or appearance in addition to expression and pose.}
\label{fig:teaser}
\end{teaserfigure} 

\maketitle
\thispagestyle{empty}
\fancyfoot{}

\section{Introduction}
\label{sec:introduction}
Hundreds of millions of images are captured and shared across the world every single day.
Many of these are portraits: single-image snapshots of a person's head.
While portrait images are abundant, the appearance and expression of the subject are fixed at the point of capture.
The ability to edit a portrait image, or to infer what the subject might look like if captured with a different pose, expression, hairstyle or even physical age, has significant relevance to applications in computer graphics, vision, and virtual reality. 

A convincing method for portrait image animation enables image manipulation \red{while preserving the likeness or identity of the subject}. 
When manipulating the pose or posture of an individual, the method should produce multi-view consistent results such that the \red{identity} is not altered during the animation.

A number of approaches have been proposed to achieve this goal.
For example, methods based on generative adversarial network (GAN) inversion manipulate the latent space of a pre-trained 2D GAN~\cite{choi2020starganv2,karras2018progressive,karras2019style,karras2020stylegan2,karras2021alias,wang2018high} to adjust expression, pose, lighting, or other attributes~\cite{harkonen2020ganspace,shen2020interpreting,tewari2020stylerig,tewari2020pie}.
This approach, however, often results in unpredictable changes to the appearance or identity with changes in viewpoint due to the lack of modeling the 3D structure of the scene.
Other methods based on 3D morphable models (3DMMs)~\cite{blanz1999morphable} predict an explicit textured mesh representation of the portrait image subject and can thus manipulate appearance through animation of the underlying mesh~\cite{DECA:Siggraph2021,deng2019accurate,kim2018deep,thies2016face2face}.
While successful in editing facial expressions, these approaches are not capable of directly editing hair or semantic attributes.
Moreover, these methods cannot usually be directly to portrait image animation since they usually require multiple images or a video sequence of a subject in order to synthesize a complete texture map.
Finally, another class of methods animates portrait images by warping an input image based on a desired pose and expression and refining the output using a neural network~\cite{Siarohin_2019_NeurIPS,ren2021pirenderer,Zakharov20}.
While these methods produce plausible results, they do not achieve the photorealistic quality of GAN-based methods, and generally have view inconsistencies for large deviations from the initial portrait image.

In this work, we address these limitations to achieve multi-view-consistent editing and animation of portrait images.
Our approach leverages recently developed 3D-aware GANs, specifically EG3D \cite{Chan2021}, 
and a 3DMM-based supervision strategy that enables fine-grain control over expression editing. 
For an input portrait image, we estimate a 3DMM using a pre-trained encoder network~\cite{DECA:Siggraph2021}.
Using the 3DMM, we edit the facial expression of the portrait image in a controllable fashion.
A main insight is that this edited image can be used to supervise the GAN to produce the same edited expression using a photometric loss.
To this end, we perform GAN inversion and fine-tuning to simultaneously embed the 3DMM-synthesized image into the latent space of the GAN and in-paint any masked out regions (e.g., the interior of the mouth).
After inversion, the 3D GAN predicts an explicit neural radiance field~\cite{mildenhall2020nerf}, enabling multi-view consistent rendering from arbitrary poses across the edited expressions.

Our approach can similarly be used to drive portrait images for video re-enactment.
We use the 3DMM parameters from the video to edit the portrait image expressions and embed each of the resulting frames into the GAN latent space.
Likewise, by leveraging the semantic editing capabilities of the GAN, we can control pose while also editing attributes such as age, hairstyle, and gender (see Fig.~\ref{fig:teaser}).

In summary, we make the following contributions:
\begin{itemize}[nosep]	
	\item We propose a method for explicit manipulation of expression and pose of a portrait image by fine-tuning the latent space of a pre-trained 3D GAN with 3DMM-based supervision.
    \item We extend this approach for video re-enactment from a portrait image and show an improvement in synthesized image quality compared to other methods.
    \item We demonstrate attribute-edited portrait image animation; that is, we exploit the latent space of the 3D GAN model in order to edit semantic attributes such as age, hairstyle, or gender in addition to explicitly controlling the expression and pose.
\end{itemize}

\paragraph{Overview of Limitations.}
Here, we mention a few limitations of our work.
First, because our method uses a pre-trained 3D GAN~\cite{Chan2021} trained on a 2D image dataset, the camera pose and head pose are entangled.
Thus, manipulating head pose is achieved with a corresponding rotation and translation of the camera, which results in movement of the background and shoulders.
While this can potentially be addressed by developing new training techniques that disentangle head and camera pose, this is out of scope of the current work, as we focus on modeling the appearance of the face and head. 
Second, we are limited by the characteristics of the FFHQ dataset~\cite{karras2019style} used to train the 3D GAN.
As a result, editing eyes is challenging since most GAN-synthesized images have open eyes that are looking directly at the camera.

\section{Related Work}
\label{sec:related}
In this section, we outline the most relevant and recent work in the area of animating portrait images. 
\begin{figure*}
    \centering
    \includegraphics[width=\textwidth]{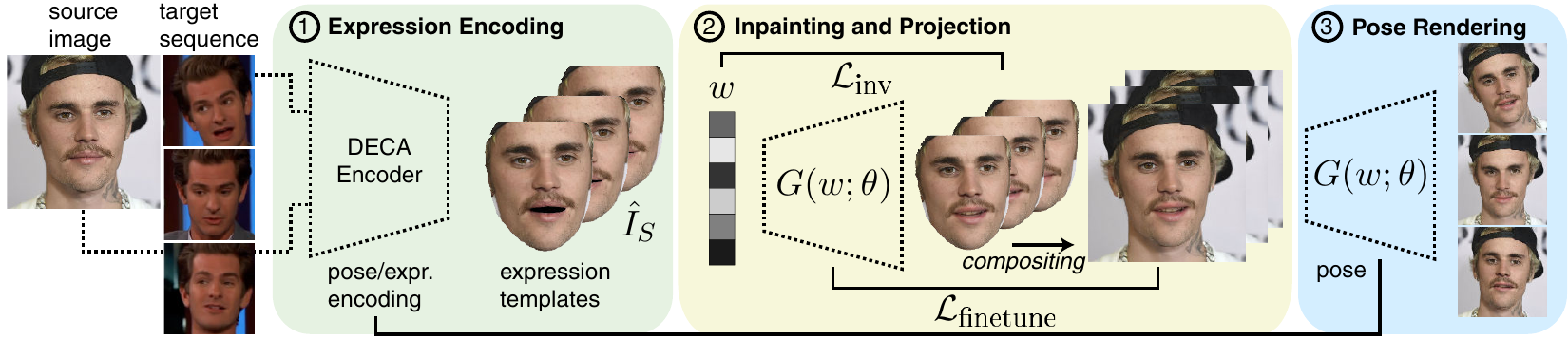}
    \caption{Overview of the proposed portrait image animation method. Given a source image and target image sequence, our method transfers pose and expression attributes to the source image. (1) We encode the expressions of the target image frames and transfer them to the source image using a 3DMM predicted with a pre-trained model~\cite{DECA:Siggraph2021}. Using the resulting 3DMM, we render expression templates, or images of the source image face with the target expressions. (2) GAN inversion is used to re-render the expression templates, which also in-paints the mouth region if necessary. The in-painted mouth is composited back onto the expression template and image background, and the result is embedded into the GAN latent space using Pivotal Tuning Inversion~\cite{roich2021pivotal}. (3) The final result is rendered by explicitly conditioning the 3D GAN with the poses of the target sequence.}
    \label{fig:method}
\end{figure*}
\paragraph{Model-based Portrait Image Editing}
Traditionally, related tasks were addressed by fitting 3D morphable models (3DMMs)~\cite{blanz1999morphable,DECA:Siggraph2021} to a 2D image or video and then editing or re-enacting this model (e.g.,~\cite{thies2016face2face,thies2019deferred}).
3DMM-based approaches, however, typically only model the face rather than other details such as hair or teeth.
Moreover, these models usually require a texture atlas built from multiple views of the subject to handle disocclusions.
A recent survey of these approaches can be found in the state-of-the-art (SoTA) report by Zollh\"ofer et al.~\shortcite{zollhofer2018state}.

\paragraph{Machine Learning--based Portrait Image and Video Editing}

Over the last few years, neural synthesis of realistic talking head sequences has become a topic of interest.
Autoencoder-type network architectures are often used with videos~\cite{pix2pix2017,kim2018deep,Lombardi:2018,wang2019fewshotvid2vid}, few-shot video clips~\cite{MarioNETte:AAAI2020,Zakharov19,wang2019fewshotvid2vid}, or single-images as input~\cite{Averbuch:2017,Shu-ECCV18,Wiles18,Geng:2018,Nagano:2018,tewari2020state,tewari2021advances,tripathy+kannala+rahtu,Siarohin_2019_NeurIPS,Zakharov20}.
These approaches can animate a single target image using the motion of a source video. Autoencoder networks, however, make it difficult to attain intuitive parametric control of the animation.
Recent approaches aim to overcome this challenge using a keypoint-driven motion flow field~\cite{wang2021facevid2vid} or a 3DMM~\cite{ren2021pirenderer} to drive the portrait image edits.
Yet, all these network architectures are oblivious to the underlying 3D structure of the content, making it challenging to produce photorealistic outputs across a range of viewpoints.
To address this problem, recent work combines neural radiance field and 3DMMs~\cite{gafni2021nerface}, but this approach requires a video showing the target person from all angles for building an editable representation.
In our work, we focus on intuitive and 3DMM-driven editing of a single target portrait image in a multi-view consistent manner. 

\paragraph{Portrait Image Editing using GANs}

SoTA portrait image editing techniques operate in the latent space of a GAN~\cite{radford2016unsupervised,karras2018progressive,karras2019style,karras2020stylegan2,karras2021alias}.
The latent space of StyleGAN, in particular, has been studied extensively and enabled image editing using linear and non-linear latent space arithmetic~\cite{Pumarola_ijcv2019,abdal2019image2stylegan,abdal2020image2stylegan++,abdal2021styleflow,harkonen2020ganspace,shen2020interpreting,zhu2020domain,Tov:2021,roich2021pivotal}; local, semantically-aware edits~\cite{Collins20}; 3DMM-driven control of StyleGAN's latent space~\cite{tewari2020stylerig,tewari2020pie}; a combination of traditional face rendering and inverse GANs~\cite{chandran2021rendering}; and free-viewpoint rendering~\cite{FreeStyleGAN2021} through GAN inversion \red{(see Xia et al. for a survey of GAN inversion techniques~\shortcite{xia2021gan})}.
StyleGAN and related architectures, however, are trained to generate 2D images, which makes it challenging to control head pose, camera perspective, or other 3D attributes in a photorealistic manner.
SoFGAN~\cite{sofgan} recently addressed some of these issues by generating multi-view consistent 3D semantic labels that are textured using a conditional StyleGAN architecture.
Yet, this approach is not multi-view consistent as it relies on a conditional 2D GAN to generate the RGB pixel values.

Recent work on 3D GANs show promise in overcoming some of these limitations, but many existing architectures do not achieve the photorealistic image quality of StyleGAN.
For example, mesh-based 3D GANs are limited in viewing angle and detail~\cite{Szabo:2019,Liao2020CVPR,shi2021lifting}; voxel-based 3D GANs are limited in their resolution due to extensive memory requirements~\cite{wu2016learning,Gadelha:2017,zhu2018visual,henzler2019platonicgan,hologan,nguyen2020blockgan,hao2021GANcraft}, 3D GANs building on implicit representation networks are computationally inefficient~\cite{graf,chan2020pi}, and 3D GANs relying on 2D CNN-based image upsampling layers often suffer from view inconsistencies~\cite{Niemeyer2020GIRAFFE,gu2021stylenerf}. 

Our work builds on the EG3D architecture~\cite{Chan2021}, a recent 3D-aware GAN that is both computationally efficient and offers multi-view consistent image quality almost on par with StyleGAN2.
We extend EG3D by exploring its latent space and capabilities of animating a single target image using either intuitive editing of attributes, such as age or gender, or transferring the facial expressions and pose from a target video. 

Note that a number of concurrently developed, but still unpublished 3D-aware GANs have recently appeared~\cite{deng2021gram,daras2021solving,xu2021volumegan,orel2021styleSDF,gu2021stylenerf,zhou2021cips3d,sun2021fenerf}.
Some of these architectures may also be suitable to animate a single target image, although exploring all of these unpublished works is beyond the scope of this paper.

\section{Inverse 3D GAN Framework}
\label{sec:methods}
In order to animate and edit an input source image, we combine 3DMM-based expression editing with latent space manipulation of a pre-trained 3D GAN.
An overview of the method is shown in Fig.~\ref{fig:method}.
Given a target video, we first leverage a 3D face reconstruction model to predict and transfer facial expressions from each target frame to the source image.
Then, we perform 3D GAN inversion on the re-expressed source images, projecting them into the latent space of the GAN.
This process automatically in-paints any unconstrained regions, such as the interior of the mouth, which are not rendered by the 3DMM.
Finally, we use the 3D GAN to re-render the expression-edited source images to match the poses of the target video frames.
Our pipeline naturally supports facial attribute editing---the source image is edited using any GAN-based attribute editing method and then re-animated following the same process as above.

\subsection{3D GAN}

We choose EG3D~\cite{Chan2021} as the pre-trained 3D GAN, as it is the current SOTA for synthesizing photo-realistic, view-consistent images. Given a randomly sampled latent code $z$, the mapping network maps $z$ to an intermediate latent code $w$. This intermediate latent code modulates the convolution kernels of a synthesis network to produce tri-plane features. To render the generated image at a desired camera pose, these feature planes are sampled and decoded into a neural radiance field for volumetric rendering. The resulting synthesized images are view-consistent, allowing the 3D GAN to robustly handle occlusion and pose shifts when animating a subject. Please see Chan et al.~\shortcite{Chan2021} for additional details on the 3D GAN.

\subsection{Inversion Framework}
Given an input source image $I_S$ and a target image $I_T$, we transfer facial expressions and head pose from $I_T$ to $I_S$ while handling occlusions and preserving the facial identity of $I_S$.

\textit{Expression Transfer.} To transfer facial expressions, we employ a pre-trained DECA model~\cite{DECA:Siggraph2021} for 3D face reconstruction. We apply DECA's encoder to $I_S$ and $I_T$ to obtain the source and target identities $(\beta_S \in \mathbb{R}^{50}, \beta_T \in \mathbb{R}^{50})$, expressions $(\theta_S \in \mathbb{R}^{50}, \theta_T \in \mathbb{R}^{50})$, and poses $(\psi_S \in \mathbb{R}^6, \psi_T \in \mathbb{R}^6)$. We then form an input $(\beta_S, \theta_T, [\psi_{S_{0:3}}, \psi_{T_{3:6}}])$ for DECA's decoder, consisting of the source's identity, target's expression, source's head pose, and target jaw pose. The decoder predicts a 3DMM that we render using textures sampled from $I_S$ to obtain the expression transferred RGB and face mask output $(\hat{I}_S, M_S)$. Although we re-use the source's head pose, swapping in the target's jaw pose causes the rendered result to have disocclusions in the mouth region when the source's mouth is closed and the target's mouth is open. We rely on the 3D GAN to in-paint these regions in a realistic manner during the inversion process. Because DECA does not handle blinking expressions, the proposed pipeline inherits this limitation and does not transfer such eye expressions. 

\textit{3D GAN Inversion.} Given the re-expressed source image and face mask $(\hat{I}_S, M_S)$ from DECA, we draw inspiration from Pivotal Tuning Inversion (PTI)~\cite{roich2021pivotal} and perform 3D GAN inversion in two stages. In the first stage, we optimize for the latent code $w$ that most closely reconstructs the re-expressed face region by minimizing the loss $\mathcal{L}_{\text{inv}}$
\begin{align}
    M_{S_{ij}} &= \begin{dcases} 1 & \hat{I}_{S_{ij}} \in \text{face} \wedge \hat{I}_{S_{ij}} \not\in \text{mouth} \\ 0 & \text{otherwise} \end{dcases}\\
    \mathcal{L}_{\text{inv}} &= \| \Phi(M_S \odot \hat{I}_S) - \Phi(M_S \odot G(w; \theta)) \|_2.
\end{align}
Here, $\Phi$ is an image feature extractor, such as VGG~\cite{simonyan2014very}, and $\theta$ are the parameters of the generator, which are frozen during this stage. We limit the optimization to the masked face region to avoid overly constraining the latent space of the generator, as in-the-wild images often contain unseen backgrounds and subjects. During this stage, the generator leverages its prior to realistically hallucinate occlusions near the mouth region.

In the second stage, we freeze the optimized latent code $w^*$ and fine-tune the generator to match the rest of $\hat{I}_S$. 
Note that we also freeze the in-painted mouth region, as we find empirically that in-painted regions tend to degenerate unless explicitly constrained throughout the optimization of the generator.
We define the loss function used to fine-tune the generator as follows.
\begin{align}
    M_{\text{mouth}_{ij}} &= \begin{dcases} 1 & \hat{I}_{S_{ij}} \in \text{mouth} \\ 0 & \text{otherwise} \end{dcases} \\
    I_{\text{GT}} &= (M_{\text{mouth}} \odot \hat{I}_S) + (1 - M_{\text{mouth}}) \odot I_S\\
    I_{\text{pred}} &= G(w^*; \theta),    \\
    \mathcal{L}_{\text{finetune}} &= \mathcal{L}_{\text{LPIPS}}(I_\text{GT}, I_\text{pred}) + \| I_\text{GT} - I_\text{pred} \|_2,  
\end{align}
$\mathcal{L}_{\text{LPIPS}}$ is a perceptual loss function~\cite{zhang2018perceptual}, $M_{\text{mouth}}$ is the mask including all pixels except the mouth region, and $\theta^*$ are the fine-tuned generator parameters.

\begin{figure}
    \centering
    \includegraphics[width=\columnwidth]{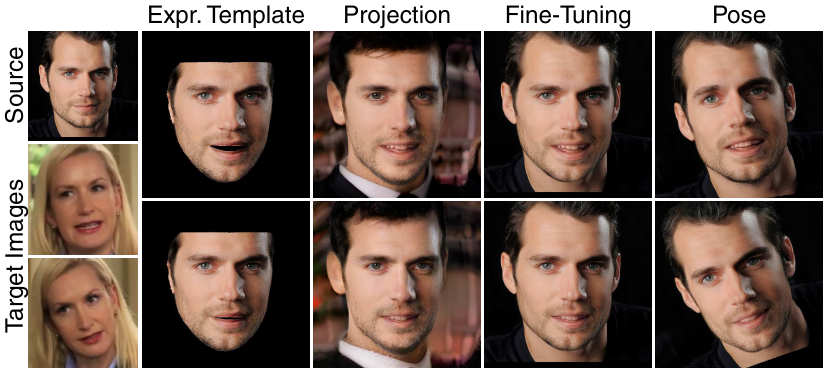}
    \caption{Visualization of intermediate outputs from the image animation pipeline.
    The expression template is produced using an estimated 3DMM. Then, the face is projected into the latent space of the 3D GAN, which also in-paints the mouth region. After finetuning, the hair and background match the source image and the pose can be manipulated directly by the 3D GAN.}
    \label{fig:pipeline}
\end{figure}

\textit{Head Pose Transfer.} One major benefit of the 3D GAN is that pose is naturally disentangled and can be easily manipulated while maintaining view consistency. After inversion, we simply re-render $I_{\textrm{pred}}$ using the 3D GAN to match the target head pose, estimated using Deep3DFace~\cite{deng2019accurate}.
The final output is the edited portrait image with an expression and pose corresponding to the target image. 

We show intermediate outputs of this process in Fig.~\ref{fig:pipeline}.
Here, the expression template is the expression-edited source image produced using the 3DMM. 
Then, after the projection step, we visualize the result of rendering $w^*$ using the 3D GAN generator.
Note that the face region closely matches the expression template and the mouth region is in-painted.
After fine-tuning the generator, regions outside the face, including the hair and background, closely match the original source image.
Finally, the pose is adjusted to match the target frames by conditioning the 3D GAN directly with the corresponding pose values.


\subsection{Video-based Face Animation} \label{sec:videoface}

For an input target video $V_T$ consisting of frames $I_{T_0}, I_{T_1}, \ldots, I_{T_N}$, we adapt the same expression transfer pipeline to each frame. We smooth the estimated poses over a window of two frames to mitigate camera jittering during head pose transfer. Although the source input is a single image, the 3D GAN naturally facilitates static image animation due to its view consistency. However, subtle flickering artifacts may still arise from the 3D GAN being unable to exactly recover the background and shoulders during inversion. To address this, we apply the re-projected face mask to obtain just the face region $F_i$ for each frame $i > 1$. We then render the first frame using pose $i$ and composite in $F_i$. Here, we leverage the view consistency of the 3D GAN to ensure that the final result is temporally consistent.

\subsection{Facial Attribute Editing}

We incorporate attribute editing into the proposed pipeline using an edit-then-fit approach.
Specifically, we adapt StyleFlow~\cite{abdal2021styleflow} to work with the pre-trained 3D GAN;
given an input $w$ and desired attribute modifications, the method predicts a new $w$ that results in an image with the desired adjustments when rendered using the 3D GAN.
This attribute-edited image is used as the input to the rest of the image animation pipeline, described previously.
 
To adapt StyleFlow for the 3D GAN, we prepare a dataset of 10,000 paired latent codes $z$ and $w$, sampled randomly from the latent space of the pre-trained 3D GAN. 
We use the 3D GAN to render the portrait images corresponding to each $w$, the Microsoft Face API~\cite{azure} to generate attribute labels for each image.
After assembling this dataset we follow Abdal et al.~\shortcite{abdal2021styleflow} to train a continuous normalizing flow~\cite{kobyzev2020normalizing} to provide an invertible mapping between $w$ and $z$, conditioned on the attributes. 

After training, we edit a real input image $I$ by performing PTI to retrieve the $w$ that best reconstructs $I$ and retrieving the image attributes using the Face API. We use the continuous normalizing flow to obtain the original $z$ that maps to the optimized $w$:
\begin{equation}
    z(t_0) = z(t_1) + \int_{t_1}^{t_0} \phi(z(t), t, \alpha; \theta) dt.
\end{equation} 
Here, $z(t_0) = z, z(t_1) = w, \alpha$ is the vector of attributes, and $\phi$ is the continuous normalizing flow network.
We modify elements of $\alpha$ corresponding to the attributes we wish to edit, forming the updated attribute vector $\alpha^*$.
Finally, we perform a forward flow to produce the latent code $w^* = z(t_1)$ that encodes the edited image $I^*$ with attributes $\alpha^*$:
\begin{equation}
    z(t_1) = z(t_0) + \int_{t_0}^{t_1} \phi(z(t), t, \alpha^*; \theta) dt.
\end{equation} 

\begin{figure*}
    \centering
    \includegraphics[width=\textwidth]{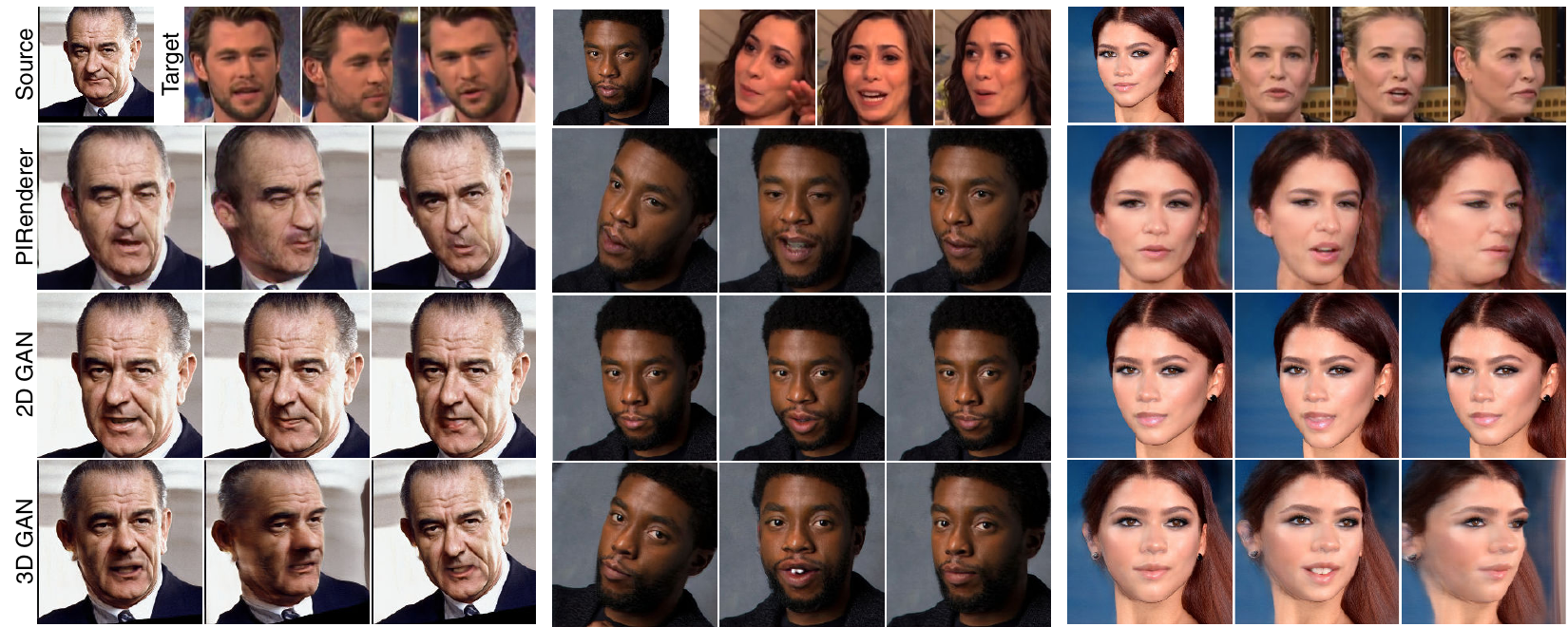}
    \caption{Portrait image animation results. We compare the proposed approach with PIRenderer~\cite{ren2021pirenderer} and the proposed approach using a pre-trained 2D GAN~\cite{karras2020stylegan2} instead of the 3D GAN. Since we cannot directly control the pose with the 2D GAN, we only edit the expressions. The proposed 3D GAN-based approach achieves better view consistency with higher output image quality compared to PIRenderer.}
    \label{fig:editing}
\end{figure*}

To animate the edited source image, we repeat the steps detailed in Sec.~\ref{sec:videoface} by synthesizing re-expressed source images using DECA~\cite{DECA:Siggraph2021}.
The re-expressed images then serve as input into the 3D GAN inversion pipeline to produce the final outputs.

\sloppy{While StyleFlow or other latent space interpolation techniques~\cite{shen2020interpreting,harkonen2020ganspace} can be used with 2D GANs to control pose via conditional attributes, the results are limited in pose range} and suffer from view inconsistencies and changes to the subject likeness during pose manipulation.
Since the 3D GAN predicts a fixed radiance field that can be rendered from any pose, the proposed method is naturally multi-view consistent and preserves the subject likeness better than the 2D approach as we demonstrate in the following section.

\section{Results}
\label{sec:experiments}
We present results of the proposed method with comparisons to other baseline methods, including PIRenderer~\cite{ren2021pirenderer}, a SoTA approach for portrait image animation, and a modified version of the proposed method using a 2D GAN.
PIRenderer uses a pre-trained 3DMM encoder to predict the expressions of the target frame and transfers them to the source image using a warping field predicted with a convolutional network.
A separate convolutional block then refines the result to produce the final output image.
The 2D GAN-based method is a straightforward adaptation of the proposed method, except that pose cannot be directly controlled. 

Overall, the 3D GAN provides far better animation quality than the 2D GAN because the pose can be explicitly controlled.
While both the 2D GAN and 3D GAN show qualitative and quantitative improvements over PIRenderer in terms of image quality, PIRenderer matches the target expressions more closely, primarily because it is not explicitly constrained by the expressiveness of the face model used for the 2D and 3D GAN.
We also show that the pose of the 2D GAN can be manipulated using StyleFlow, but that manipulating the pose using the 3D GAN produces significantly better multi-view consistency and \red{identity preservation} across edits.

\subsection{Portrait Image Animation}
We evaluate the proposed method for portrait image animation and compare against the PIRenderer baseline~\cite{ren2021pirenderer} and the proposed approach using a pre-trained 2D GAN~\cite{karras2020stylegan2}.
\paragraph{Implementation Details.}
The 3D GAN is adapted directly from Chan et al.~\cite{Chan2021} with pre-training on the Flicker-Faces-HQ (FFHQ) dataset~\cite{karras2019style} and an output resolution of $512\times 512$ pixels.
We adapt DECA to perform the expression encoding with some modifications to the original codebase~\cite{DECA:Siggraph2021} to \red{render the predicted 3DMM face using a high-resolution texture map extracted from the input source image.} 
After rendering out the face masks using DECA, we optimize $w$ for the first frame in the image sequence for 2,000 steps using the Adam optimizer~\cite{kingma2014adam}. 
For subsequent frames, we initialize $w$ with the value from the previous frame and run the optimization for an additional 100 steps. 
We fine-tune the generator for 10,000 steps using a batch size of 1.
The whole procedure takes roughly two hours for a sequence of 100 target images running on an NVIDIA A6000 GPU.
To improve temporal consistency in the final result, we re-render the first frame of the sequence with the pose of each subsequent frame. 
For all other frames, we extract the rendered face and composite it back onto the first-frame image with the same pose. This procedure removes any jitter or variation in the background pixels between frames, and the face extraction can be implemented automatically by re-projecting the 3DMM face mask onto the target pose using the depth maps predicted by the 3D GAN.
We will make all code publicly available prior to publication.

\paragraph{Qualitative Results.}
We show qualitative comparisons of the proposed method, PIRenderer, and the approach using the 2D GAN in Fig.~\ref{fig:editing} with additional results in the supplemental.
PIRenderer shows good control over the expressions, including the mouth and eyes of the source image.
However, the output images tend to be lower fidelity than results from the 2D or 3D GAN, with blurrier details and patchy artifacts where the network attempts to in-paint missing details.
Moreover, since there is no explicit 3D view consistency, the \red{identity} of the source image may shift as the pose changes.
Qualitatively, the 2D and 3D GAN results are similar, but since there is no native mechanism for manipulating the 2D GAN pose, only the expression is modulated. Since the 3DMM framework cannot be used to synthesize eye blinks and both the 2D GAN and 3D GAN are trained on FFHQ~\cite{karras2019style}, a dataset in which few subjects are blinking, the proposed framework cannot adjust the eye position or blink state. 
Still, the 3D GAN achieves a good match to the target pose with similar expressions while reproducing fine details of the input image.
Importantly, the 3D GAN also preserves the \red{identity} of the input image during pose adjustments.
\red{Note that the pretrained and 2D and 3D GANs require slightly different image crops, and so we crop the images according to their respective implementations (the 2D GAN crop is somewhat more centered on the entire head, see Fig.~\ref{fig:editing}).}

\begin{table}[ht]
    \centering
    \resizebox{\columnwidth}{!}{
    \begin{tabular}{lcc|cc}
    \toprule
        & FID$\downarrow$  & ID $\uparrow$ & APD$\downarrow$ & AED$\downarrow$ \\\midrule
        PIRenderer (w/o eyes, w/o pose) & 53.916            & -              & 0.250          & 0.437 \\
        PIRenderer (w/o pose)           & 53.959            & -              & 0.247          & 0.386 \\
        PIRenderer (w/o eyes)           & 63.844            & 0.694          & 0.039          & 0.424 \\
        PIRenderer                      & 64.379            & 0.700          & 0.040          & \textbf{0.373} \\\midrule
        2D GAN (w/o pose)               & 17.812            & -              & 0.246          & 0.434 \\\midrule
        3D GAN (w/o pose)               & \textbf{16.504}   & -              & 0.246          & 0.433 \\
        3D GAN                          & 31.176            & \textbf{0.733} & \textbf{0.030} & 0.433 \\
    \bottomrule
\end{tabular}}
\caption{Quantitative evaluation of portrait image animation. We compare our approach using the pre-trained 3D GAN of Chan et al.~\shortcite{Chan2021} to our approach using a 2D GAN~\cite{karras2020stylegan2} and PIRenderer~\cite{ren2021pirenderer}. The 3D GAN achieves the best performance in terms of Fr\`echet Inception Distance (FID), identity consistency (ID), avg.\ pose distance (APD), while PIRenderer achieves the best avg.\ expression distance (AED).}
    \label{tab:metrics}
\end{table}

\begin{table}[ht]
    \centering
    \resizebox{\columnwidth}{!}{
    \begin{tabular}{ccccccc}
    \toprule
    \multicolumn{4}{c|}{PIRenderer} & 2D GAN & \multicolumn{2}{|c}{3D GAN} \\
    w/o eyes, w/o pose & w/o pose & w/o eyes & \multicolumn{1}{c|}{full} & \multicolumn{1}{c|}{w/o pose} & w/o pose & full \\\midrule
    0.652 & 0.622 & 0.476 & 0.451 & 0.891 & 0.896 & 0.590 \\\bottomrule
\end{tabular}}
\caption{\red{Evaluation of identity preservation compared to the source image. The provided ArcFace similarity score~\cite{deng2018arcface} has values between -1 and 1 (greater value means more similar) and is calculated by comparing the rendered images to the input source image. Scores improve when removing pose or eye modeling, but the 3D GAN produces the best results compared to relevant variants of PIRenderer or the 2D GAN.}}
    \label{tab:id_source}
\end{table}

\paragraph{Quantitative Results}
We observe similar trends in the quantitative results, reported in Table~\ref{tab:metrics}.
These results are calculated over a dataset of 500 images from CelebA-HQ~\cite{karras2018progressive} and 500 target video sequences sampled from VoxCeleb~\cite{Nagrani19,Chung18b,Nagrani17}. 
We evaluate using the Fr\`echet Inception Distance (FID), average pose distance (APD), average expression distance (AED), and an identity consistency metric (ID).

FID is computed between the original 500 source images and 500 samples of the edited source images, averaged over the 100 possible realizations of these sets of images, sampled without replacement.
For each video clip, the APD and AED are computed between the modified source images and the target images.
The APD is given as root mean squared error between the estimated pose encodings from a pre-trained Deep3DFace model~\cite{deng2019accurate} for the modified source image and the target image sequence.
The AED is defined similarly, except the metric is calculated between the expression encodings predicted by the pre-trained model. 
Finally, we also compute an identity consistency metric using a pre-trained Arcface model~\cite{deng2018arcface}.
For this metric, we randomly sample 2,000 poses from the CelebA-HQ dataset and randomly apply two poses to two different frames from each video clip for a total of 1000 image pairs. 
We calculate the cosine distance of the predicted embedding for each pair and report the average result. 
\red{To evaluate how well the identity is preserved compared to the source embedding, we randomly sample two rendered frames corresponding to each of the 500 source images and compute the identity consistency metric between these samples and their source images. 
Results are shown in Table~\ref{tab:id_source}}

\begin{figure*}
    \centering
    \includegraphics[width=\textwidth]{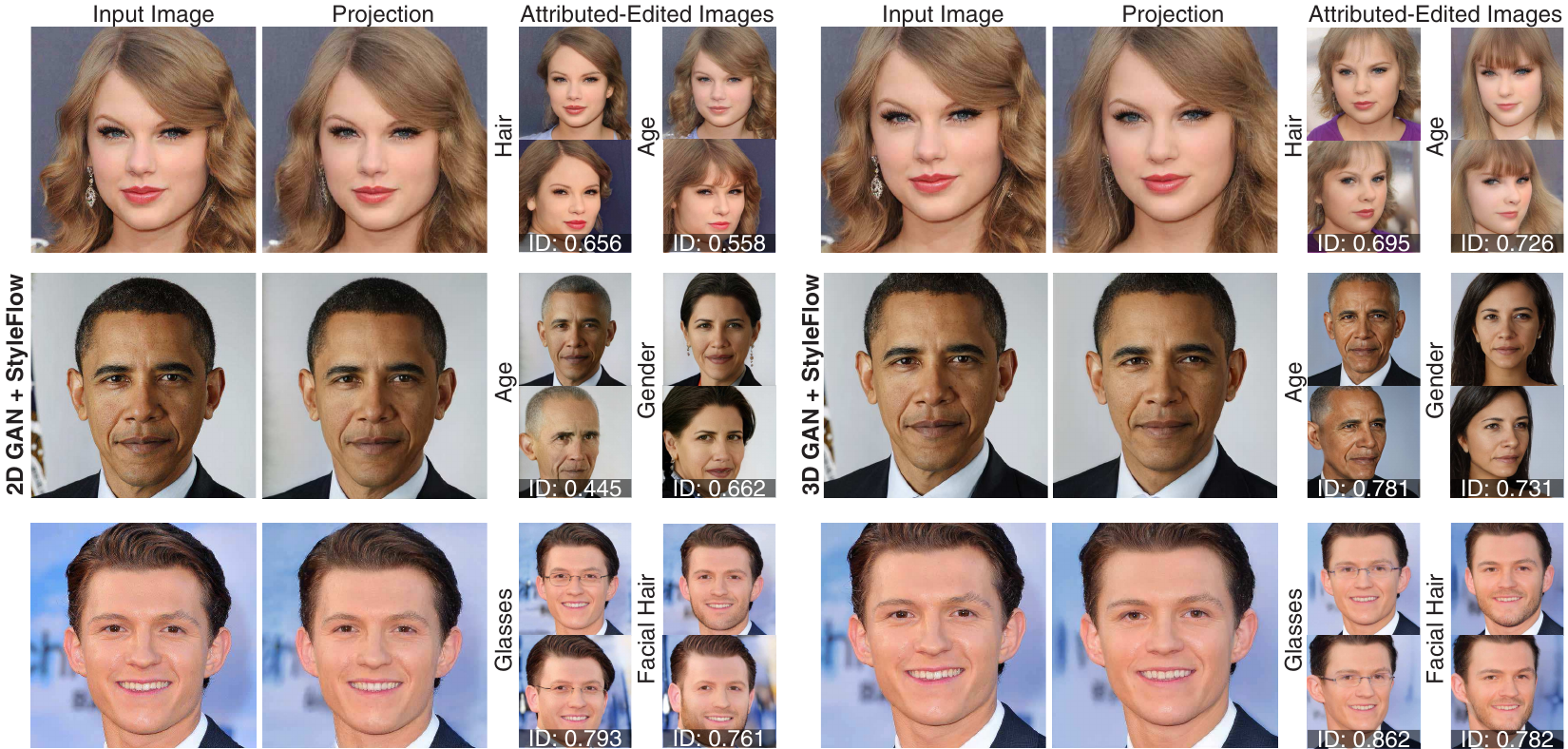}
    \caption{Attribute editing results. We compare attribute editing and portrait image manipulation using the 2D GAN~\cite{karras2020stylegan2} and the 3D GAN~\cite{Chan2021}. We perform attribute edits using StyleFlow~\cite{abdal2021styleflow} and adapt this method for the 3D GAN. Qualitatively both GANs achieve similarly expressive edits, but the 3D GAN demonstrates greater multi-view consistency when adjusting the pose after editing, as shown by the 3D GAN's higher ID similarity scores~\cite{deng2018arcface}. The two GANs expect slightly different input crops and pose is controlled for the 2D GAN via StyleFlow editing.}
    \label{fig:attribute}
\end{figure*}

The proposed method using the 3D GAN achieves the best accuracy in terms of FID score, average pose distance, and identity consistency (see Table~\ref{tab:metrics}).
We hypothesize that the limited ability of our model to capture eye motion (e.g., blinking) results in a penalty to the expression distance metric.
A similar drop in performance is observed for a variant of PIRenderer with fixed eye expressions (see ``w/o eyes'' in Table~\ref{tab:metrics}), which we achieve by fixing the eye vertices of the 3DMM model to those of the first frame in the target image sequence.
Finally, since the 2D GAN does not explicitly model pose, we observe that its performance is similar to a variant of the 3D GAN with fixed poses or PIRenderer where poses and eye expressions are fixed. Note that methods with fixed pose (``w/o pose'') have improved FID scores because they are not animated away from the initial poses of the ground truth image dataset.
\red{Similarly, variants of the methods with fixed pose or eyes have improved identity consistency for the same reason (Table~\ref{tab:id_source}), though the 3D GAN shows the best consistency among the relevant baseline variants.}

\begin{table}[t]
    \centering
    \resizebox{\columnwidth}{!}{
    \begin{tabular}{lcccccc}
    \toprule
        & Pose Only & + Age & + Facial Hair & + Gender & + Glasses & + Hair  \\
        2D GAN & 0.652 & 0.637 & 0.638 & 0.652 & 0.653 & 0.686 \\
        3D GAN & 0.790 & 0.797 & 0.790 & 0.793 & 0.794 & 0.801 \\
    \bottomrule
\end{tabular}}
\caption{Quantitative evaluation of multi-view consistency and face identity preservation via \red{ArcFace cosine similarity scores~\cite{deng2018arcface} -- a score of -1 and 1 indicate maximum dissimilarity or similarity, respectively.}}
    \label{tab:attribute}
\end{table}

\subsection{Attribute Editing}

\paragraph{Implementation Details}
We use a pre-trained StyleGAN2 model for the 2D GAN baseline and a pre-trained EG3D model for our 3D GAN. We re-train a StyleFlow model for each method following the original paper. To generate each training dataset, we randomly sample 10,000 $w$ latent vectors with a truncation factor of 0.7 to reduce the number of low quality outlier faces. Both StyleFlow models are trained for 20 epochs using a batch size of 5. To perform edits for an input image, we first perform Pivotal Tuning Inversion~\cite{roich2021pivotal} with 500 $w$ optimization steps and only $100$ generator fine-tuning steps with locality regularization to minimize latent space distribution shift.

\paragraph{Qualitative Results}
Fig.~\ref{fig:attribute} compares attribute editing results between the 2D GAN and 3D GAN using StyleFlow to control hair, age, glasses, gender, and facial hair. Qualitatively, both the 2D GAN and the 3D GAN can reconstruct the input image faithfully and exhibit similarly reasonable editing results for frontal facing subjects. \red{In both cases, we observe some shift in the identity, e.g., when modifying hairstyle which seems to be a limitation of the StyleFlow editing directions and perhaps the expressiveness of the GAN latent spaces.}  However, the 3D GAN is significantly more view-consistent, as the 2D GAN undergoes a large degree of identity shift when the pitch and yaw attributes are shifted. 

\paragraph{Quantitative Results}

We evaluate multi-view consistency and face identity preservation by measuring ArcFace cosine similarity.  As shown by Fig.~\ref{fig:attribute}, editing with the 3D GAN produces higher similarity scores than with the 2D GAN. Table~\ref{tab:attribute} demonstrates that this trend also holds for generated subjects. We randomly sample 128 subjects from each GAN's latent space, apply an attribute edit to each subject, and render out two views for each subject. These poses are uniformly sampled from yaw angles between $[-30^\circ, 30^\circ]$ and pitch angles between $[-20^\circ, 20^\circ]$.

\section{Discussion}
\label{sec:discussion}
We have presented a method for controllable portrait image animation based on 3D GAN inversion. 
The proposed approach combines explicit control over expressions afforded by 3DMMs with the abilities of emerging 3D GANs to edit semantic information, in-paint missing details, and explicitly control the pose of rendered portrait images.
Overall, leveraging a 3D GAN for portrait image animation significantly improves image quality over non-generative approaches and multi-view consistency over 2D GAN methods. 

Still, there are a number of limitations to the proposed method.
Firstly, the method is constrained by the underlying face model. A direct consequence is that the current implementation does not support rendering eye blinks or changes to the eye position.
We also acknowledge limitations of the 3D GAN used in the method; the FFHQ dataset used to train the 3D GAN is not calibrated to represent a balanced cross-section of human appearances. As a result, portrait animation and attribute editing performance is somewhat dependent on the input image appearance. 
These limitations could be alleviated by using PCA-based editing techniques~\cite{harkonen2020ganspace} to synthesize blinking textures, more sophisticated face tracking models, or improving the GAN training procedure to disentangle the subject expression, allowing explicit control similar to what is currently possible for the pose.

Overall, this work takes important steps towards improving the quality of portrait image animation and the utility of pre-trained GANs for controllable image manipulation.
We envision that this work could have relevance to applications across graphics and interactive techniques, including for social media, virtual telecommunications, automated talking head synthesis, and realistic re-dubbing for television and film.

\paragraph{Societal Impact.}
The proposed method enables the synthesis and animation of photorealistic images; recent advances in this area have enabled new and incredible applications, but also have potential for misuse.
We condemn the use of this work for ill-intentioned purposes, not limited to misinformation or impersonation, and support the research of methods to thwart such efforts.
We refer the reader to Tewari et al.~\shortcite{tewari2020state} for an extended discussion of methods for detecting and analyzing deepfake videos.

\begin{acks}
    This project was supported in part by a PECASE by the ARO, NSF award 1839974, Stanford HAI, and a Samsung GRO.
\end{acks}

\bibliographystyle{ACM-Reference-Format}
\bibliography{references}


\begin{thebibliography}{79}


\ifx \showCODEN    \undefined \def \showCODEN     #1{\unskip}     \fi
\ifx \showDOI      \undefined \def \showDOI       #1{#1}\fi
\ifx \showISBNx    \undefined \def \showISBNx     #1{\unskip}     \fi
\ifx \showISBNxiii \undefined \def \showISBNxiii  #1{\unskip}     \fi
\ifx \showISSN     \undefined \def \showISSN      #1{\unskip}     \fi
\ifx \showLCCN     \undefined \def \showLCCN      #1{\unskip}     \fi
\ifx \shownote     \undefined \def \shownote      #1{#1}          \fi
\ifx \showarticletitle \undefined \def \showarticletitle #1{#1}   \fi
\ifx \showURL      \undefined \def \showURL       {\relax}        \fi
\providecommand\bibfield[2]{#2}
\providecommand\bibinfo[2]{#2}
\providecommand\natexlab[1]{#1}
\providecommand\showeprint[2][]{arXiv:#2}

\bibitem[\protect\citeauthoryear{Abdal, Qin, and Wonka}{Abdal
  et~al\mbox{.}}{2019}]%
        {abdal2019image2stylegan}
\bibfield{author}{\bibinfo{person}{Rameen Abdal}, \bibinfo{person}{Yipeng Qin},
  {and} \bibinfo{person}{Peter Wonka}.} \bibinfo{year}{2019}\natexlab{}.
\newblock \showarticletitle{Image2StyleGAN: How to Embed Images into the
  StyleGAN Latent Space?}. In \bibinfo{booktitle}{\emph{IEEE International
  Conference on Computer Vision (ICCV)}}.
\newblock


\bibitem[\protect\citeauthoryear{Abdal, Qin, and Wonka}{Abdal
  et~al\mbox{.}}{2020}]%
        {abdal2020image2stylegan++}
\bibfield{author}{\bibinfo{person}{Rameen Abdal}, \bibinfo{person}{Yipeng Qin},
  {and} \bibinfo{person}{Peter Wonka}.} \bibinfo{year}{2020}\natexlab{}.
\newblock \showarticletitle{Image2StyleGAN++: How to Edit the Embedded
  Images?}. In \bibinfo{booktitle}{\emph{IEEE Conference on Computer Vision and
  Pattern Recognition (CVPR)}}.
\newblock


\bibitem[\protect\citeauthoryear{Abdal, Zhu, Mitra, and Wonka}{Abdal
  et~al\mbox{.}}{2021}]%
        {abdal2021styleflow}
\bibfield{author}{\bibinfo{person}{Rameen Abdal}, \bibinfo{person}{Peihao Zhu},
  \bibinfo{person}{Niloy~J Mitra}, {and} \bibinfo{person}{Peter Wonka}.}
  \bibinfo{year}{2021}\natexlab{}.
\newblock \showarticletitle{StyleFlow: Attribute-conditioned Exploration of
  StyleGAN-Generated Images using Conditional Continuous Normalizing Flows}.
\newblock \bibinfo{journal}{\emph{ACM Transactions on Graphics (ToG)}}
  \bibinfo{volume}{40}, \bibinfo{number}{3} (\bibinfo{year}{2021}).
\newblock


\bibitem[\protect\citeauthoryear{Averbuch-Elor, Cohen-Or, Kopf, and
  Cohen}{Averbuch-Elor et~al\mbox{.}}{2017}]%
        {Averbuch:2017}
\bibfield{author}{\bibinfo{person}{Hadar Averbuch-Elor},
  \bibinfo{person}{Daniel Cohen-Or}, \bibinfo{person}{Johannes Kopf}, {and}
  \bibinfo{person}{Michael~F. Cohen}.} \bibinfo{year}{2017}\natexlab{}.
\newblock \showarticletitle{Bringing Portraits to Life}.
\newblock \bibinfo{journal}{\emph{ACM Transactions on Graphics (ToG)}}
  \bibinfo{volume}{36}, \bibinfo{number}{6} (\bibinfo{year}{2017}).
\newblock


\bibitem[\protect\citeauthoryear{Blanz and Vetter}{Blanz and Vetter}{1999}]%
        {blanz1999morphable}
\bibfield{author}{\bibinfo{person}{Volker Blanz} {and} \bibinfo{person}{Thomas
  Vetter}.} \bibinfo{year}{1999}\natexlab{}.
\newblock \showarticletitle{A Morphable Model for the Synthesis of 3D Faces}.
  In \bibinfo{booktitle}{\emph{Proceedings of SIGGRAPH}}.
\newblock


\bibitem[\protect\citeauthoryear{Chan, Lin, Chan, Nagano, Pan, Mello, Gallo,
  Guibas, Tremblay, Khamis, Karras, and Wetzstein}{Chan et~al\mbox{.}}{2022}]%
        {Chan2021}
\bibfield{author}{\bibinfo{person}{Eric~R. Chan}, \bibinfo{person}{Connor~Z.
  Lin}, \bibinfo{person}{Matthew~A. Chan}, \bibinfo{person}{Koki Nagano},
  \bibinfo{person}{Boxiao Pan}, \bibinfo{person}{Shalini~De Mello},
  \bibinfo{person}{Orazio Gallo}, \bibinfo{person}{Leonidas Guibas},
  \bibinfo{person}{Jonathan Tremblay}, \bibinfo{person}{Sameh Khamis},
  \bibinfo{person}{Tero Karras}, {and} \bibinfo{person}{Gordon Wetzstein}.}
  \bibinfo{year}{2022}\natexlab{}.
\newblock \showarticletitle{Efficient Geometry-aware {3D} Generative
  Adversarial Networks}. In \bibinfo{booktitle}{\emph{IEEE Conference on
  Computer Vision and Pattern Recognition (CVPR)}}.
\newblock


\bibitem[\protect\citeauthoryear{Chan, Monteiro, Kellnhofer, Wu, and
  Wetzstein}{Chan et~al\mbox{.}}{2021}]%
        {chan2020pi}
\bibfield{author}{\bibinfo{person}{Eric~R Chan}, \bibinfo{person}{Marco
  Monteiro}, \bibinfo{person}{Petr Kellnhofer}, \bibinfo{person}{Jiajun Wu},
  {and} \bibinfo{person}{Gordon Wetzstein}.} \bibinfo{year}{2021}\natexlab{}.
\newblock \showarticletitle{{pi-GAN}: {P}eriodic Implicit Generative
  Adversarial Networks for {3D}-Aware Image Synthesis}. In
  \bibinfo{booktitle}{\emph{IEEE Conference on Computer Vision and Pattern
  Recognition (CVPR)}}.
\newblock


\bibitem[\protect\citeauthoryear{Chandran, Winberg, Zoss, Riviere, Gross,
  Gotardo, and Bradley}{Chandran et~al\mbox{.}}{2021}]%
        {chandran2021rendering}
\bibfield{author}{\bibinfo{person}{Prashanth Chandran},
  \bibinfo{person}{Sebastian Winberg}, \bibinfo{person}{Gaspard Zoss},
  \bibinfo{person}{J{\'e}r{\'e}my Riviere}, \bibinfo{person}{Markus Gross},
  \bibinfo{person}{Paulo Gotardo}, {and} \bibinfo{person}{Derek Bradley}.}
  \bibinfo{year}{2021}\natexlab{}.
\newblock \showarticletitle{Rendering with Style: Combining Traditional and
  Neural Approaches for High-Quality Face Rendering}.
\newblock \bibinfo{journal}{\emph{ACM Transactions on Graphics (ToG)}}
  \bibinfo{volume}{40}, \bibinfo{number}{6} (\bibinfo{year}{2021}).
\newblock


\bibitem[\protect\citeauthoryear{Chen, Liu, Xie, Chen, Su, and Jingyi}{Chen
  et~al\mbox{.}}{2021}]%
        {sofgan}
\bibfield{author}{\bibinfo{person}{Anpei Chen}, \bibinfo{person}{Ruiyang Liu},
  \bibinfo{person}{Ling Xie}, \bibinfo{person}{Zhang Chen},
  \bibinfo{person}{Hao Su}, {and} \bibinfo{person}{Yu Jingyi}.}
  \bibinfo{year}{2021}\natexlab{}.
\newblock \showarticletitle{SofGAN: A Portrait Image Generator with Dynamic
  Styling}.
\newblock \bibinfo{journal}{\emph{ACM Trans. Graph.}} \bibinfo{volume}{41},
  \bibinfo{number}{1} (\bibinfo{year}{2021}).
\newblock


\bibitem[\protect\citeauthoryear{Choi, Uh, Yoo, and Ha}{Choi
  et~al\mbox{.}}{2020}]%
        {choi2020starganv2}
\bibfield{author}{\bibinfo{person}{Yunjey Choi}, \bibinfo{person}{Youngjung
  Uh}, \bibinfo{person}{Jaejun Yoo}, {and} \bibinfo{person}{Jung-Woo Ha}.}
  \bibinfo{year}{2020}\natexlab{}.
\newblock \showarticletitle{StarGAN v2: Diverse Image Synthesis for Multiple
  Domains}. In \bibinfo{booktitle}{\emph{Proceedings of the IEEE Conference on
  Computer Vision and Pattern Recognition}}.
\newblock


\bibitem[\protect\citeauthoryear{Chung, Nagrani, and Zisserman}{Chung
  et~al\mbox{.}}{2018}]%
        {Chung18b}
\bibfield{author}{\bibinfo{person}{J.~S. Chung}, \bibinfo{person}{A. Nagrani},
  {and} \bibinfo{person}{A. Zisserman}.} \bibinfo{year}{2018}\natexlab{}.
\newblock \showarticletitle{VoxCeleb2: Deep Speaker Recognition}. In
  \bibinfo{booktitle}{\emph{Interspeech}}.
\newblock


\bibitem[\protect\citeauthoryear{Collins, Bala, Price, and
  S{\"u}sstrunk}{Collins et~al\mbox{.}}{2020}]%
        {Collins20}
\bibfield{author}{\bibinfo{person}{Edo Collins}, \bibinfo{person}{Raja Bala},
  \bibinfo{person}{Bob Price}, {and} \bibinfo{person}{Sabine S{\"u}sstrunk}.}
  \bibinfo{year}{2020}\natexlab{}.
\newblock \showarticletitle{Editing in Style: Uncovering the Local Semantics of
  {GANs}}. In \bibinfo{booktitle}{\emph{IEEE Conference on Computer Vision and
  Pattern Recognition (CVPR)}}.
\newblock


\bibitem[\protect\citeauthoryear{Daras, Chu, Kumar, Lagun, and Dimakis}{Daras
  et~al\mbox{.}}{2021}]%
        {daras2021solving}
\bibfield{author}{\bibinfo{person}{Giannis Daras}, \bibinfo{person}{Wen-Sheng
  Chu}, \bibinfo{person}{Abhishek Kumar}, \bibinfo{person}{Dmitry Lagun}, {and}
  \bibinfo{person}{Alexandros~G. Dimakis}.} \bibinfo{year}{2021}\natexlab{}.
\newblock \bibinfo{title}{Solving Inverse Problems with NerfGANs}.
\newblock
\newblock
\showeprint[arxiv]{2112.09061}~[cs.CV]


\bibitem[\protect\citeauthoryear{Deng, Guo, Niannan, and Zafeiriou}{Deng
  et~al\mbox{.}}{2019a}]%
        {deng2018arcface}
\bibfield{author}{\bibinfo{person}{Jiankang Deng}, \bibinfo{person}{Jia Guo},
  \bibinfo{person}{Xue Niannan}, {and} \bibinfo{person}{Stefanos Zafeiriou}.}
  \bibinfo{year}{2019}\natexlab{a}.
\newblock \showarticletitle{ArcFace: Additive Angular Margin Loss for Deep Face
  Recognition}. In \bibinfo{booktitle}{\emph{CVPR}}.
\newblock


\bibitem[\protect\citeauthoryear{Deng, Yang, Xiang, and Tong}{Deng
  et~al\mbox{.}}{2021}]%
        {deng2021gram}
\bibfield{author}{\bibinfo{person}{Yu Deng}, \bibinfo{person}{Jiaolong Yang},
  \bibinfo{person}{Jianfeng Xiang}, {and} \bibinfo{person}{Xin Tong}.}
  \bibinfo{year}{2021}\natexlab{}.
\newblock \showarticletitle{GRAM: Generative Radiance Manifolds for 3D-Aware
  Image Generation}. In \bibinfo{booktitle}{\emph{arXiv}}.
\newblock


\bibitem[\protect\citeauthoryear{Deng, Yang, Xu, Chen, Jia, and Tong}{Deng
  et~al\mbox{.}}{2019b}]%
        {deng2019accurate}
\bibfield{author}{\bibinfo{person}{Yu Deng}, \bibinfo{person}{Jiaolong Yang},
  \bibinfo{person}{Sicheng Xu}, \bibinfo{person}{Dong Chen},
  \bibinfo{person}{Yunde Jia}, {and} \bibinfo{person}{Xin Tong}.}
  \bibinfo{year}{2019}\natexlab{b}.
\newblock \showarticletitle{Accurate 3D Face Reconstruction with
  Weakly-Supervised Learning: From Single Image to Image Set}. In
  \bibinfo{booktitle}{\emph{IEEE Conference on Computer Vision and Pattern
  Recognition (CVPR)}}.
\newblock


\bibitem[\protect\citeauthoryear{Feng, Feng, Black, and Bolkart}{Feng
  et~al\mbox{.}}{2021}]%
        {DECA:Siggraph2021}
\bibfield{author}{\bibinfo{person}{Yao Feng}, \bibinfo{person}{Haiwen Feng},
  \bibinfo{person}{Michael~J. Black}, {and} \bibinfo{person}{Timo Bolkart}.}
  \bibinfo{year}{2021}\natexlab{}.
\newblock \showarticletitle{Learning an Animatable Detailed {3D} Face Model
  from In-The-Wild Images}.
\newblock \bibinfo{journal}{\emph{ACM Transactions on Graphics (SIGGRAPH)}}
  \bibinfo{volume}{40}, \bibinfo{number}{8}.
\newblock


\bibitem[\protect\citeauthoryear{Gadelha, Maji, and Wang}{Gadelha
  et~al\mbox{.}}{2017}]%
        {Gadelha:2017}
\bibfield{author}{\bibinfo{person}{Matheus Gadelha}, \bibinfo{person}{Subhransu
  Maji}, {and} \bibinfo{person}{Rui Wang}.} \bibinfo{year}{2017}\natexlab{}.
\newblock \showarticletitle{{3D} Shape Induction from {2D} Views of Multiple
  Objects}. In \bibinfo{booktitle}{\emph{International Conference on 3D
  Vision}}.
\newblock


\bibitem[\protect\citeauthoryear{Gafni, Thies, Zollh{\"o}fer, and
  Nie{\ss}ner}{Gafni et~al\mbox{.}}{2021}]%
        {gafni2021nerface}
\bibfield{author}{\bibinfo{person}{Guy Gafni}, \bibinfo{person}{Justus Thies},
  \bibinfo{person}{Michael Zollh{\"o}fer}, {and} \bibinfo{person}{Matthias
  Nie{\ss}ner}.} \bibinfo{year}{2021}\natexlab{}.
\newblock \showarticletitle{Dynamic Neural Radiance Fields for Monocular 4D
  Facial Avatar Reconstruction}. In \bibinfo{booktitle}{\emph{IEEE Conference
  on Computer Vision and Pattern Recognition (CVPR)}}.
\newblock


\bibitem[\protect\citeauthoryear{Geng, Shao, Zheng, Weng, and Zhou}{Geng
  et~al\mbox{.}}{2018}]%
        {Geng:2018}
\bibfield{author}{\bibinfo{person}{Jiahao Geng}, \bibinfo{person}{Tianjia
  Shao}, \bibinfo{person}{Youyi Zheng}, \bibinfo{person}{Yanlin Weng}, {and}
  \bibinfo{person}{Kun Zhou}.} \bibinfo{year}{2018}\natexlab{}.
\newblock \showarticletitle{Warp-Guided GANs for Single-Photo Facial
  Animation}.
\newblock \bibinfo{journal}{\emph{ACM Transactions on Graphics (ToG)}}
  \bibinfo{volume}{37}, \bibinfo{number}{6} (\bibinfo{year}{2018}).
\newblock


\bibitem[\protect\citeauthoryear{Gu, Liu, Wang, and Theobalt}{Gu
  et~al\mbox{.}}{2021}]%
        {gu2021stylenerf}
\bibfield{author}{\bibinfo{person}{Jiatao Gu}, \bibinfo{person}{Lingjie Liu},
  \bibinfo{person}{Peng Wang}, {and} \bibinfo{person}{Christian Theobalt}.}
  \bibinfo{year}{2021}\natexlab{}.
\newblock \showarticletitle{{StyleNeRF}: {A} Style-based {3D}-Aware Generator
  for High-resolution Image Synthesis}.
\newblock \bibinfo{journal}{\emph{arXiv preprint arXiv:2110.08985}}
  (\bibinfo{year}{2021}).
\newblock


\bibitem[\protect\citeauthoryear{Ha, Kersner, Kim, Seo, and Kim}{Ha
  et~al\mbox{.}}{2020}]%
        {MarioNETte:AAAI2020}
\bibfield{author}{\bibinfo{person}{Sungjoo Ha}, \bibinfo{person}{Martin
  Kersner}, \bibinfo{person}{Beomsu Kim}, \bibinfo{person}{Seokjun Seo}, {and}
  \bibinfo{person}{Dongyoung Kim}.} \bibinfo{year}{2020}\natexlab{}.
\newblock \showarticletitle{MarioNETte: Few-shot Face Reenactment Preserving
  Identity of Unseen Targets}. In \bibinfo{booktitle}{\emph{Proceedings of the
  AAAI Conference on Artificial Intelligence}}.
\newblock


\bibitem[\protect\citeauthoryear{Hao, Mallya, Belongie, and Liu}{Hao
  et~al\mbox{.}}{2021}]%
        {hao2021GANcraft}
\bibfield{author}{\bibinfo{person}{Zekun Hao}, \bibinfo{person}{Arun Mallya},
  \bibinfo{person}{Serge Belongie}, {and} \bibinfo{person}{Ming-Yu Liu}.}
  \bibinfo{year}{2021}\natexlab{}.
\newblock \showarticletitle{{GANcraft}: {U}nsupervised {3D} Neural Rendering of
  Minecraft Worlds}. In \bibinfo{booktitle}{\emph{IEEE International Conference
  on Computer Vision (ICCV)}}.
\newblock


\bibitem[\protect\citeauthoryear{H\"ark\"onen, Hertzmann, Lehtinen, and
  Paris}{H\"ark\"onen et~al\mbox{.}}{2020}]%
        {harkonen2020ganspace}
\bibfield{author}{\bibinfo{person}{Erik H\"ark\"onen}, \bibinfo{person}{Aaron
  Hertzmann}, \bibinfo{person}{Jaakko Lehtinen}, {and} \bibinfo{person}{Sylvain
  Paris}.} \bibinfo{year}{2020}\natexlab{}.
\newblock \showarticletitle{GANSpace: Discovering Interpretable GAN Controls}.
  In \bibinfo{booktitle}{\emph{Advances in Neural Information Processing
  Systems (NeurIPS)}}.
\newblock


\bibitem[\protect\citeauthoryear{Henzler, Mitra, and Ritschel}{Henzler
  et~al\mbox{.}}{2019}]%
        {henzler2019platonicgan}
\bibfield{author}{\bibinfo{person}{Philipp Henzler}, \bibinfo{person}{Niloy~J
  Mitra}, {and} \bibinfo{person}{Tobias Ritschel}.}
  \bibinfo{year}{2019}\natexlab{}.
\newblock \showarticletitle{Escaping {P}lato's Cave: {3D} Shape From
  Adversarial Rendering}. In \bibinfo{booktitle}{\emph{IEEE International
  Conference on Computer Vision (ICCV)}}.
\newblock


\bibitem[\protect\citeauthoryear{Isola, Zhu, Zhou, and Efros}{Isola
  et~al\mbox{.}}{2017}]%
        {pix2pix2017}
\bibfield{author}{\bibinfo{person}{Phillip Isola}, \bibinfo{person}{Jun-Yan
  Zhu}, \bibinfo{person}{Tinghui Zhou}, {and} \bibinfo{person}{Alexei~A
  Efros}.} \bibinfo{year}{2017}\natexlab{}.
\newblock \showarticletitle{Image-to-Image Translation with Conditional
  Adversarial Networks}.
\newblock \bibinfo{journal}{\emph{CVPR}} (\bibinfo{year}{2017}).
\newblock


\bibitem[\protect\citeauthoryear{Karras, Aila, Laine, and Lehtinen}{Karras
  et~al\mbox{.}}{2018}]%
        {karras2018progressive}
\bibfield{author}{\bibinfo{person}{Tero Karras}, \bibinfo{person}{Timo Aila},
  \bibinfo{person}{Samuli Laine}, {and} \bibinfo{person}{Jaakko Lehtinen}.}
  \bibinfo{year}{2018}\natexlab{}.
\newblock \showarticletitle{Progressive Growing of {GANs} for Improved Quality,
  Stability, and Variation}. In \bibinfo{booktitle}{\emph{International
  Conference on Learning Representations (ICLR)}}.
\newblock


\bibitem[\protect\citeauthoryear{Karras, Aittala, Laine, H{\"a}rk{\"o}nen,
  Hellsten, Lehtinen, and Aila}{Karras et~al\mbox{.}}{2021}]%
        {karras2021alias}
\bibfield{author}{\bibinfo{person}{Tero Karras}, \bibinfo{person}{Miika
  Aittala}, \bibinfo{person}{Samuli Laine}, \bibinfo{person}{Erik
  H{\"a}rk{\"o}nen}, \bibinfo{person}{Janne Hellsten}, \bibinfo{person}{Jaakko
  Lehtinen}, {and} \bibinfo{person}{Timo Aila}.}
  \bibinfo{year}{2021}\natexlab{}.
\newblock \showarticletitle{Alias-free Generative Adversarial Networks}.
\newblock \bibinfo{journal}{\emph{Advances in Neural Information Processing
  Systems}}  \bibinfo{volume}{34} (\bibinfo{year}{2021}).
\newblock


\bibitem[\protect\citeauthoryear{Karras, Laine, and Aila}{Karras
  et~al\mbox{.}}{2019}]%
        {karras2019style}
\bibfield{author}{\bibinfo{person}{Tero Karras}, \bibinfo{person}{Samuli
  Laine}, {and} \bibinfo{person}{Timo Aila}.} \bibinfo{year}{2019}\natexlab{}.
\newblock \showarticletitle{A Style-Based Generator Architecture for Generative
  Adversarial Networks}. In \bibinfo{booktitle}{\emph{IEEE Conference on
  Computer Vision and Pattern Recognition (CVPR)}}.
\newblock


\bibitem[\protect\citeauthoryear{Karras, Laine, Aittala, Hellsten, Lehtinen,
  and Aila}{Karras et~al\mbox{.}}{2020}]%
        {karras2020stylegan2}
\bibfield{author}{\bibinfo{person}{Tero Karras}, \bibinfo{person}{Samuli
  Laine}, \bibinfo{person}{Miika Aittala}, \bibinfo{person}{Janne Hellsten},
  \bibinfo{person}{Jaakko Lehtinen}, {and} \bibinfo{person}{Timo Aila}.}
  \bibinfo{year}{2020}\natexlab{}.
\newblock \showarticletitle{Analyzing and Improving the Image Quality of
  {StyleGAN}}. In \bibinfo{booktitle}{\emph{IEEE Conference on Computer Vision
  and Pattern Recognition (CVPR)}}.
\newblock


\bibitem[\protect\citeauthoryear{Kim, Garrido, Tewari, Xu, Thies, Nie{\ss}ner,
  P{\'e}rez, Richardt, Zoll{\"o}fer, and Theobalt}{Kim et~al\mbox{.}}{2018}]%
        {kim2018deep}
\bibfield{author}{\bibinfo{person}{Hyeongwoo Kim}, \bibinfo{person}{Pablo
  Garrido}, \bibinfo{person}{Ayush Tewari}, \bibinfo{person}{Weipeng Xu},
  \bibinfo{person}{Justus Thies}, \bibinfo{person}{Matthias Nie{\ss}ner},
  \bibinfo{person}{Patrick P{\'e}rez}, \bibinfo{person}{Christian Richardt},
  \bibinfo{person}{Michael Zoll{\"o}fer}, {and} \bibinfo{person}{Christian
  Theobalt}.} \bibinfo{year}{2018}\natexlab{}.
\newblock \showarticletitle{Deep Video Portraits}.
\newblock \bibinfo{journal}{\emph{ACM Transactions on Graphics (TOG)}}
  \bibinfo{volume}{37}, \bibinfo{number}{4} (\bibinfo{year}{2018}).
\newblock


\bibitem[\protect\citeauthoryear{Kingma and Ba}{Kingma and Ba}{2014}]%
        {kingma2014adam}
\bibfield{author}{\bibinfo{person}{Diederik~P Kingma} {and}
  \bibinfo{person}{Jimmy Ba}.} \bibinfo{year}{2014}\natexlab{}.
\newblock \showarticletitle{Adam: A Method for Stochastic Optimization}.
\newblock \bibinfo{journal}{\emph{International Conference on Learning
  Representations (ICLR)}}.
\newblock


\bibitem[\protect\citeauthoryear{Kobyzev, Prince, and Brubaker}{Kobyzev
  et~al\mbox{.}}{2020}]%
        {kobyzev2020normalizing}
\bibfield{author}{\bibinfo{person}{Ivan Kobyzev}, \bibinfo{person}{Simon
  Prince}, {and} \bibinfo{person}{Marcus Brubaker}.}
  \bibinfo{year}{2020}\natexlab{}.
\newblock \showarticletitle{Normalizing Flows: An Introduction and Review of
  Current Methods}.
\newblock \bibinfo{journal}{\emph{IEEE Transactions on Pattern Analysis and
  Machine Intelligence (TPAMI)}} (\bibinfo{year}{2020}).
\newblock


\bibitem[\protect\citeauthoryear{Leimk\"uhler and Drettakis}{Leimk\"uhler and
  Drettakis}{2021}]%
        {FreeStyleGAN2021}
\bibfield{author}{\bibinfo{person}{Thomas Leimk\"uhler} {and}
  \bibinfo{person}{George Drettakis}.} \bibinfo{year}{2021}\natexlab{}.
\newblock \showarticletitle{FreeStyleGAN: Free-view Editable Portrait Rendering
  with the Camera Manifold}.
\newblock  \bibinfo{volume}{40}, \bibinfo{number}{6} (\bibinfo{year}{2021}).
\newblock


\bibitem[\protect\citeauthoryear{Liao, Schwarz, Mescheder, and Geiger}{Liao
  et~al\mbox{.}}{2020}]%
        {Liao2020CVPR}
\bibfield{author}{\bibinfo{person}{Yiyi Liao}, \bibinfo{person}{Katja Schwarz},
  \bibinfo{person}{Lars Mescheder}, {and} \bibinfo{person}{Andreas Geiger}.}
  \bibinfo{year}{2020}\natexlab{}.
\newblock \showarticletitle{Towards Unsupervised Learning of Generative Models
  for {3D} Controllable Image Synthesis}. In \bibinfo{booktitle}{\emph{IEEE
  Conference on Computer Vision and Pattern Recognition (CVPR)}}.
\newblock


\bibitem[\protect\citeauthoryear{Lombardi, Saragih, Simon, and Sheikh}{Lombardi
  et~al\mbox{.}}{2018}]%
        {Lombardi:2018}
\bibfield{author}{\bibinfo{person}{Stephen Lombardi}, \bibinfo{person}{Jason
  Saragih}, \bibinfo{person}{Tomas Simon}, {and} \bibinfo{person}{Yaser
  Sheikh}.} \bibinfo{year}{2018}\natexlab{}.
\newblock \showarticletitle{Deep Appearance Models for Face Rendering}.
\newblock \bibinfo{journal}{\emph{ACM Trans. Graph.}} \bibinfo{volume}{37},
  \bibinfo{number}{4}, Article \bibinfo{articleno}{68} (\bibinfo{date}{July}
  \bibinfo{year}{2018}), \bibinfo{numpages}{13}~pages.
\newblock
\showISSN{0730-0301}


\bibitem[\protect\citeauthoryear{Microsoft}{Microsoft}{2020}]%
        {azure}
\bibfield{author}{\bibinfo{person}{Microsoft}.}
  \bibinfo{year}{2020}\natexlab{}.
\newblock \bibinfo{title}{Azure Face}.
\newblock
\newblock
\urldef\tempurl%
\url{https://azure.microsoft.com/en-in/services/cognitive-services/face/}
\showURL{%
\tempurl}


\bibitem[\protect\citeauthoryear{Mildenhall, Srinivasan, Tancik, Barron,
  Ramamoorthi, and Ng}{Mildenhall et~al\mbox{.}}{2020}]%
        {mildenhall2020nerf}
\bibfield{author}{\bibinfo{person}{Ben Mildenhall}, \bibinfo{person}{Pratul~P
  Srinivasan}, \bibinfo{person}{Matthew Tancik}, \bibinfo{person}{Jonathan~T
  Barron}, \bibinfo{person}{Ravi Ramamoorthi}, {and} \bibinfo{person}{Ren Ng}.}
  \bibinfo{year}{2020}\natexlab{}.
\newblock \showarticletitle{{NeRF}: {R}epresenting Scenes as Neural Radiance
  Fields for View Synthesis}. In \bibinfo{booktitle}{\emph{European Conference
  on Computer Vision (ECCV)}}.
\newblock


\bibitem[\protect\citeauthoryear{Nagano, Seo, Xing, Wei, Li, Saito, Agarwal,
  Fursund, and Li}{Nagano et~al\mbox{.}}{2018}]%
        {Nagano:2018}
\bibfield{author}{\bibinfo{person}{Koki Nagano}, \bibinfo{person}{Jaewoo Seo},
  \bibinfo{person}{Jun Xing}, \bibinfo{person}{Lingyu Wei},
  \bibinfo{person}{Zimo Li}, \bibinfo{person}{Shunsuke Saito},
  \bibinfo{person}{Aviral Agarwal}, \bibinfo{person}{Jens Fursund}, {and}
  \bibinfo{person}{Hao Li}.} \bibinfo{year}{2018}\natexlab{}.
\newblock \showarticletitle{PaGAN: Real-Time Avatars Using Dynamic Textures}.
\newblock \bibinfo{journal}{\emph{ACM Transactions on Graphics (ToG)}}
  \bibinfo{volume}{37}, \bibinfo{number}{6} (\bibinfo{year}{2018}).
\newblock


\bibitem[\protect\citeauthoryear{Nagrani, Chung, Xie, and Zisserman}{Nagrani
  et~al\mbox{.}}{2019}]%
        {Nagrani19}
\bibfield{author}{\bibinfo{person}{Arsha Nagrani}, \bibinfo{person}{Joon~Son
  Chung}, \bibinfo{person}{Weidi Xie}, {and} \bibinfo{person}{Andrew
  Zisserman}.} \bibinfo{year}{2019}\natexlab{}.
\newblock \showarticletitle{VoxCeleb: Large-scale Speaker Verification in the
  Wild}.
\newblock \bibinfo{journal}{\emph{Computer Science and Language}}
  (\bibinfo{year}{2019}).
\newblock


\bibitem[\protect\citeauthoryear{Nagrani, Chung, and Zisserman}{Nagrani
  et~al\mbox{.}}{2017}]%
        {Nagrani17}
\bibfield{author}{\bibinfo{person}{A. Nagrani}, \bibinfo{person}{J.~S. Chung},
  {and} \bibinfo{person}{A. Zisserman}.} \bibinfo{year}{2017}\natexlab{}.
\newblock \showarticletitle{VoxCeleb: A Large-scale Speaker Identification
  Dataset}. In \bibinfo{booktitle}{\emph{Interspeech}}.
\newblock


\bibitem[\protect\citeauthoryear{Nguyen-Phuoc, Li, Theis, Richardt, and
  Yang}{Nguyen-Phuoc et~al\mbox{.}}{2019}]%
        {hologan}
\bibfield{author}{\bibinfo{person}{Thu Nguyen-Phuoc}, \bibinfo{person}{Chuan
  Li}, \bibinfo{person}{Lucas Theis}, \bibinfo{person}{Christian Richardt},
  {and} \bibinfo{person}{Yong-Liang Yang}.} \bibinfo{year}{2019}\natexlab{}.
\newblock \showarticletitle{{HoloGAN}: {U}nsupervised Learning of {3D}
  Representations from Natural Images}. In \bibinfo{booktitle}{\emph{IEEE
  International Conference on Computer Vision (ICCV)}}.
\newblock


\bibitem[\protect\citeauthoryear{Nguyen-Phuoc, Richardt, Mai, Yang, and
  Mitra}{Nguyen-Phuoc et~al\mbox{.}}{2020}]%
        {nguyen2020blockgan}
\bibfield{author}{\bibinfo{person}{Thu Nguyen-Phuoc},
  \bibinfo{person}{Christian Richardt}, \bibinfo{person}{Long Mai},
  \bibinfo{person}{Yong-Liang Yang}, {and} \bibinfo{person}{Niloy Mitra}.}
  \bibinfo{year}{2020}\natexlab{}.
\newblock \showarticletitle{{BlockGAN}: Learning {3D} Object-aware Scene
  Representations from Unlabelled Images}. In
  \bibinfo{booktitle}{\emph{Advances in Neural Information Processing Systems
  (NeurIPS)}}.
\newblock


\bibitem[\protect\citeauthoryear{Niemeyer and Geiger}{Niemeyer and
  Geiger}{2021}]%
        {Niemeyer2020GIRAFFE}
\bibfield{author}{\bibinfo{person}{Michael Niemeyer} {and}
  \bibinfo{person}{Andreas Geiger}.} \bibinfo{year}{2021}\natexlab{}.
\newblock \showarticletitle{{GIRAFFE}: {R}epresenting Scenes as Compositional
  Generative Neural Feature Fields}. In \bibinfo{booktitle}{\emph{IEEE
  Conference on Computer Vision and Pattern Recognition (CVPR)}}.
\newblock


\bibitem[\protect\citeauthoryear{Or-El, Luo, Shan, Shechtman, Park, and
  Kemelmacher-Shlizerman}{Or-El et~al\mbox{.}}{2021}]%
        {orel2021styleSDF}
\bibfield{author}{\bibinfo{person}{Roy Or-El}, \bibinfo{person}{Xuan Luo},
  \bibinfo{person}{Mengyi Shan}, \bibinfo{person}{Eli Shechtman},
  \bibinfo{person}{Jeong~Joon Park}, {and} \bibinfo{person}{Ira
  Kemelmacher-Shlizerman}.} \bibinfo{year}{2021}\natexlab{}.
\newblock \showarticletitle{StyleSDF: High-Resolution 3D-Consistent Image and
  Geometry Generation}.
\newblock \bibinfo{journal}{\emph{arXiv preprint arXiv:2112.11427}}
  (\bibinfo{year}{2021}).
\newblock


\bibitem[\protect\citeauthoryear{Pumarola, Agudo, Martinez, Sanfeliu, and
  Moreno-Noguer}{Pumarola et~al\mbox{.}}{2019}]%
        {Pumarola_ijcv2019}
\bibfield{author}{\bibinfo{person}{A. Pumarola}, \bibinfo{person}{A. Agudo},
  \bibinfo{person}{A.M. Martinez}, \bibinfo{person}{A. Sanfeliu}, {and}
  \bibinfo{person}{F. Moreno-Noguer}.} \bibinfo{year}{2019}\natexlab{}.
\newblock \showarticletitle{GANimation: One-Shot Anatomically Consistent Facial
  Animation}.
\newblock  (\bibinfo{year}{2019}).
\newblock


\bibitem[\protect\citeauthoryear{Radford, Metz, and Chintala}{Radford
  et~al\mbox{.}}{2016}]%
        {radford2016unsupervised}
\bibfield{author}{\bibinfo{person}{Alec Radford}, \bibinfo{person}{Luke Metz},
  {and} \bibinfo{person}{Soumith Chintala}.} \bibinfo{year}{2016}\natexlab{}.
\newblock \showarticletitle{Unsupervised Representation Learning with Deep
  Convolutional Generative Adversarial Networks}. In
  \bibinfo{booktitle}{\emph{International Conference on Learning
  Representations (ICLR)}}.
\newblock


\bibitem[\protect\citeauthoryear{Ren, Li, Chen, Li, and Liu}{Ren
  et~al\mbox{.}}{2021}]%
        {ren2021pirenderer}
\bibfield{author}{\bibinfo{person}{Yurui Ren}, \bibinfo{person}{Ge Li},
  \bibinfo{person}{Yuanqi Chen}, \bibinfo{person}{Thomas~H Li}, {and}
  \bibinfo{person}{Shan Liu}.} \bibinfo{year}{2021}\natexlab{}.
\newblock \showarticletitle{PIRenderer: Controllable Portrait Image Generation
  via Semantic Neural Rendering}. In \bibinfo{booktitle}{\emph{IEEE
  International Conference on Computer Vision (ICCV)}}.
\newblock


\bibitem[\protect\citeauthoryear{Roich, Mokady, Bermano, and Cohen-Or}{Roich
  et~al\mbox{.}}{2021}]%
        {roich2021pivotal}
\bibfield{author}{\bibinfo{person}{Daniel Roich}, \bibinfo{person}{Ron Mokady},
  \bibinfo{person}{Amit~H Bermano}, {and} \bibinfo{person}{Daniel Cohen-Or}.}
  \bibinfo{year}{2021}\natexlab{}.
\newblock \showarticletitle{Pivotal Tuning for Latent-based Editing of Real
  Images}.
\newblock \bibinfo{journal}{\emph{arXiv preprint arXiv:2106.05744}}
  (\bibinfo{year}{2021}).
\newblock


\bibitem[\protect\citeauthoryear{Schwarz, Liao, Niemeyer, and Geiger}{Schwarz
  et~al\mbox{.}}{2020}]%
        {graf}
\bibfield{author}{\bibinfo{person}{Katja Schwarz}, \bibinfo{person}{Yiyi Liao},
  \bibinfo{person}{Michael Niemeyer}, {and} \bibinfo{person}{Andreas Geiger}.}
  \bibinfo{year}{2020}\natexlab{}.
\newblock \showarticletitle{{GRAF}: {G}enerative Radiance Fields for {3D}-Aware
  Image Synthesis}. In \bibinfo{booktitle}{\emph{Advances in Neural Information
  Processing Systems (NeurIPS)}}.
\newblock


\bibitem[\protect\citeauthoryear{Shen, Gu, Tang, and Zhou}{Shen
  et~al\mbox{.}}{2020}]%
        {shen2020interpreting}
\bibfield{author}{\bibinfo{person}{Yujun Shen}, \bibinfo{person}{Jinjin Gu},
  \bibinfo{person}{Xiaoou Tang}, {and} \bibinfo{person}{Bolei Zhou}.}
  \bibinfo{year}{2020}\natexlab{}.
\newblock \showarticletitle{Interpreting the Latent Space of GANs for Semantic
  Face Editing}. In \bibinfo{booktitle}{\emph{IEEE Conference on Computer
  Vision and Pattern Recognition (CVPR)}}.
\newblock


\bibitem[\protect\citeauthoryear{Shi, Aggarwal, and Jain}{Shi
  et~al\mbox{.}}{2021}]%
        {shi2021lifting}
\bibfield{author}{\bibinfo{person}{Yichun Shi}, \bibinfo{person}{Divyansh
  Aggarwal}, {and} \bibinfo{person}{Anil~K Jain}.}
  \bibinfo{year}{2021}\natexlab{}.
\newblock \showarticletitle{Lifting {2D} StyleGAN for {3D}-Aware Face
  Generation}. In \bibinfo{booktitle}{\emph{IEEE Conference on Computer Vision
  and Pattern Recognition (CVPR)}}.
\newblock


\bibitem[\protect\citeauthoryear{Shu, Sahasrabudhe, Guler, Samaras, Paragios,
  and Kokkinos}{Shu et~al\mbox{.}}{2018}]%
        {Shu-ECCV18}
\bibfield{author}{\bibinfo{person}{Zhixin Shu}, \bibinfo{person}{Mihir
  Sahasrabudhe}, \bibinfo{person}{Riza~Alp Guler}, \bibinfo{person}{Dimitris
  Samaras}, \bibinfo{person}{Nikos Paragios}, {and} \bibinfo{person}{Iasonas
  Kokkinos}.} \bibinfo{year}{2018}\natexlab{}.
\newblock \showarticletitle{Deforming Autoencoders: Unsupervised Disentangling
  of Shape and Appearance}. In \bibinfo{booktitle}{\emph{European Conference on
  Computer Vision}}.
\newblock


\bibitem[\protect\citeauthoryear{Siarohin, Lathuilière, Tulyakov, Ricci, and
  Sebe}{Siarohin et~al\mbox{.}}{2019}]%
        {Siarohin_2019_NeurIPS}
\bibfield{author}{\bibinfo{person}{Aliaksandr Siarohin},
  \bibinfo{person}{Stéphane Lathuilière}, \bibinfo{person}{Sergey Tulyakov},
  \bibinfo{person}{Elisa Ricci}, {and} \bibinfo{person}{Nicu Sebe}.}
  \bibinfo{year}{2019}\natexlab{}.
\newblock \showarticletitle{First Order Motion Model for Image Animation}. In
  \bibinfo{booktitle}{\emph{Conference on Neural Information Processing Systems
  (NeurIPS)}}.
\newblock


\bibitem[\protect\citeauthoryear{Simonyan and Zisserman}{Simonyan and
  Zisserman}{2014}]%
        {simonyan2014very}
\bibfield{author}{\bibinfo{person}{Karen Simonyan} {and}
  \bibinfo{person}{Andrew Zisserman}.} \bibinfo{year}{2014}\natexlab{}.
\newblock \showarticletitle{Very Deep Convolutional Networks for Large-scale
  Image Recognition}.
\newblock \bibinfo{journal}{\emph{arXiv preprint arXiv:1409.1556}}
  (\bibinfo{year}{2014}).
\newblock


\bibitem[\protect\citeauthoryear{Sun, Wang, Zhang, Li, Zhang, Liu, and
  Wang}{Sun et~al\mbox{.}}{2021}]%
        {sun2021fenerf}
\bibfield{author}{\bibinfo{person}{Jingxiang Sun}, \bibinfo{person}{Xuan Wang},
  \bibinfo{person}{Yong Zhang}, \bibinfo{person}{Xiaoyu Li},
  \bibinfo{person}{Qi Zhang}, \bibinfo{person}{Yebin Liu}, {and}
  \bibinfo{person}{Jue Wang}.} \bibinfo{year}{2021}\natexlab{}.
\newblock \bibinfo{title}{FENeRF: Face Editing in Neural Radiance Fields}.
\newblock
\newblock
\showeprint[arxiv]{2111.15490}~[cs.CV]


\bibitem[\protect\citeauthoryear{Szab\'o, Meishvili, and Favaro}{Szab\'o
  et~al\mbox{.}}{2019}]%
        {Szabo:2019}
\bibfield{author}{\bibinfo{person}{Attila Szab\'o}, \bibinfo{person}{Givi
  Meishvili}, {and} \bibinfo{person}{Paolo Favaro}.}
  \bibinfo{year}{2019}\natexlab{}.
\newblock \showarticletitle{Unsupervised Generative {3D} Shape Learning from
  Natural Images}.
\newblock \bibinfo{journal}{\emph{arXiv preprint arXiv:1910.00287}}
  (\bibinfo{year}{2019}).
\newblock


\bibitem[\protect\citeauthoryear{Tewari, Elgharib, Bharaj, Bernard, Seidel,
  P{\'e}rez, Z{\"o}llhofer, and Theobalt}{Tewari et~al\mbox{.}}{2020a}]%
        {tewari2020stylerig}
\bibfield{author}{\bibinfo{person}{Ayush Tewari}, \bibinfo{person}{Mohamed
  Elgharib}, \bibinfo{person}{Gaurav Bharaj}, \bibinfo{person}{Florian
  Bernard}, \bibinfo{person}{Hans-Peter Seidel}, \bibinfo{person}{Patrick
  P{\'e}rez}, \bibinfo{person}{Michael Z{\"o}llhofer}, {and}
  \bibinfo{person}{Christian Theobalt}.} \bibinfo{year}{2020}\natexlab{a}.
\newblock \showarticletitle{StyleRig: Rigging StyleGAN for 3D Control over
  Portrait Images, CVPR 2020}. In \bibinfo{booktitle}{\emph{IEEE Conference on
  Computer Vision and Pattern Recognition (CVPR)}}.
\newblock


\bibitem[\protect\citeauthoryear{Tewari, Elgharib, BR, Bernard, Seidel,
  P{\'e}rez, Z{\"o}llhofer, and Theobalt}{Tewari et~al\mbox{.}}{2020b}]%
        {tewari2020pie}
\bibfield{author}{\bibinfo{person}{Ayush Tewari}, \bibinfo{person}{Mohamed
  Elgharib}, \bibinfo{person}{Mallikarjun BR}, \bibinfo{person}{Florian
  Bernard}, \bibinfo{person}{Hans-Peter Seidel}, \bibinfo{person}{Patrick
  P{\'e}rez}, \bibinfo{person}{Michael Z{\"o}llhofer}, {and}
  \bibinfo{person}{Christian Theobalt}.} \bibinfo{year}{2020}\natexlab{b}.
\newblock \showarticletitle{PIE: Portrait Image Embedding for Semantic
  Control}.
\newblock \bibinfo{journal}{\emph{ACM Transactions on Graphics (Proceedings
  SIGGRAPH Asia)}} \bibinfo{volume}{39}, \bibinfo{number}{6}.
\newblock
\urldef\tempurl%
\url{https://doi.org/10.1145/3414685.3417803}
\showDOI{\tempurl}


\bibitem[\protect\citeauthoryear{Tewari, Fried, Thies, Sitzmann, Lombardi,
  Sunkavalli, Martin-Brualla, Simon, Saragih, Nie{\ss}ner,
  et~al\mbox{.}}{Tewari et~al\mbox{.}}{2020c}]%
        {tewari2020state}
\bibfield{author}{\bibinfo{person}{Ayush Tewari}, \bibinfo{person}{Ohad Fried},
  \bibinfo{person}{Justus Thies}, \bibinfo{person}{Vincent Sitzmann},
  \bibinfo{person}{Stephen Lombardi}, \bibinfo{person}{Kalyan Sunkavalli},
  \bibinfo{person}{Ricardo Martin-Brualla}, \bibinfo{person}{Tomas Simon},
  \bibinfo{person}{Jason Saragih}, \bibinfo{person}{Matthias Nie{\ss}ner},
  {et~al\mbox{.}}} \bibinfo{year}{2020}\natexlab{c}.
\newblock \showarticletitle{State of the Art on Neural Rendering}.
\newblock \bibinfo{journal}{\emph{Eurographics Association}}
  (\bibinfo{year}{2020}).
\newblock


\bibitem[\protect\citeauthoryear{Tewari, Thies, Mildenhall, Srinivasan,
  Tretschk, Wang, Lassner, Sitzmann, Martin-Brualla, Lombardi, Simon, Theobalt,
  Niessner, Barron, Wetzstein, Zollhoefer, and Golyanik}{Tewari
  et~al\mbox{.}}{2021}]%
        {tewari2021advances}
\bibfield{author}{\bibinfo{person}{Ayush Tewari}, \bibinfo{person}{Justus
  Thies}, \bibinfo{person}{Ben Mildenhall}, \bibinfo{person}{Pratul
  Srinivasan}, \bibinfo{person}{Edgar Tretschk}, \bibinfo{person}{Yifan Wang},
  \bibinfo{person}{Christoph Lassner}, \bibinfo{person}{Vincent Sitzmann},
  \bibinfo{person}{Ricardo Martin-Brualla}, \bibinfo{person}{Stephen Lombardi},
  \bibinfo{person}{Tomas Simon}, \bibinfo{person}{Christian Theobalt},
  \bibinfo{person}{Matthias Niessner}, \bibinfo{person}{Jonathan~T. Barron},
  \bibinfo{person}{Gordon Wetzstein}, \bibinfo{person}{Michael Zollhoefer},
  {and} \bibinfo{person}{Vladislav Golyanik}.} \bibinfo{year}{2021}\natexlab{}.
\newblock \showarticletitle{Advances in Neural Rendering}.
\newblock \bibinfo{journal}{\emph{arXiv preprint arXiv:2111.05849}}
  (\bibinfo{year}{2021}).
\newblock


\bibitem[\protect\citeauthoryear{Thies, Zollh{\"o}fer, and Nie{\ss}ner}{Thies
  et~al\mbox{.}}{2019}]%
        {thies2019deferred}
\bibfield{author}{\bibinfo{person}{Justus Thies}, \bibinfo{person}{Michael
  Zollh{\"o}fer}, {and} \bibinfo{person}{Matthias Nie{\ss}ner}.}
  \bibinfo{year}{2019}\natexlab{}.
\newblock \showarticletitle{Deferred Neural Rendering: Image Synthesis using
  Neural Textures}.
\newblock \bibinfo{journal}{\emph{ACM Transactions on Graphics (ToG)}}
  \bibinfo{volume}{38}, \bibinfo{number}{4} (\bibinfo{year}{2019}).
\newblock


\bibitem[\protect\citeauthoryear{Thies, Zollhofer, Stamminger, Theobalt, and
  Nie{\ss}ner}{Thies et~al\mbox{.}}{2016}]%
        {thies2016face2face}
\bibfield{author}{\bibinfo{person}{Justus Thies}, \bibinfo{person}{Michael
  Zollhofer}, \bibinfo{person}{Marc Stamminger}, \bibinfo{person}{Christian
  Theobalt}, {and} \bibinfo{person}{Matthias Nie{\ss}ner}.}
  \bibinfo{year}{2016}\natexlab{}.
\newblock \showarticletitle{Face2Face: Real-time Face Capture and Reenactment
  of RGB Videos}. In \bibinfo{booktitle}{\emph{IEEE Conference on Computer
  Vision and Pattern Recognition (CVPR)}}.
\newblock


\bibitem[\protect\citeauthoryear{Tov, Alaluf, Nitzan, Patashnik, and
  Cohen-Or}{Tov et~al\mbox{.}}{2021}]%
        {Tov:2021}
\bibfield{author}{\bibinfo{person}{Omer Tov}, \bibinfo{person}{Yuval Alaluf},
  \bibinfo{person}{Yotam Nitzan}, \bibinfo{person}{Or Patashnik}, {and}
  \bibinfo{person}{Daniel Cohen-Or}.} \bibinfo{year}{2021}\natexlab{}.
\newblock \showarticletitle{Designing an Encoder for StyleGAN Image
  Manipulation}.
\newblock \bibinfo{journal}{\emph{ACM Transactions on Graphics (ToG)}}
  \bibinfo{volume}{40}, \bibinfo{number}{4} (\bibinfo{year}{2021}).
\newblock


\bibitem[\protect\citeauthoryear{Tripathy, Kannala, and Rahtu}{Tripathy
  et~al\mbox{.}}{2019}]%
        {tripathy+kannala+rahtu}
\bibfield{author}{\bibinfo{person}{Soumya Tripathy}, \bibinfo{person}{Juho
  Kannala}, {and} \bibinfo{person}{Esa Rahtu}.}
  \bibinfo{year}{2019}\natexlab{}.
\newblock \showarticletitle{ICface: Interpretable and Controllable Face
  Reenactment Using GANs}.
\newblock \bibinfo{journal}{\emph{arXiv preprint arXiv:1904.01909}}
  (\bibinfo{year}{2019}).
\newblock


\bibitem[\protect\citeauthoryear{Wang, Liu, Tao, Liu, Kautz, and
  Catanzaro}{Wang et~al\mbox{.}}{2019}]%
        {wang2019fewshotvid2vid}
\bibfield{author}{\bibinfo{person}{Ting-Chun Wang}, \bibinfo{person}{Ming-Yu
  Liu}, \bibinfo{person}{Andrew Tao}, \bibinfo{person}{Guilin Liu},
  \bibinfo{person}{Jan Kautz}, {and} \bibinfo{person}{Bryan Catanzaro}.}
  \bibinfo{year}{2019}\natexlab{}.
\newblock \showarticletitle{Few-shot Video-to-Video Synthesis}. In
  \bibinfo{booktitle}{\emph{Conference on Neural Information Processing Systems
  (NeurIPS)}}.
\newblock


\bibitem[\protect\citeauthoryear{Wang, Liu, Zhu, Tao, Kautz, and
  Catanzaro}{Wang et~al\mbox{.}}{2018}]%
        {wang2018high}
\bibfield{author}{\bibinfo{person}{Ting-Chun Wang}, \bibinfo{person}{Ming-Yu
  Liu}, \bibinfo{person}{Jun-Yan Zhu}, \bibinfo{person}{Andrew Tao},
  \bibinfo{person}{Jan Kautz}, {and} \bibinfo{person}{Bryan Catanzaro}.}
  \bibinfo{year}{2018}\natexlab{}.
\newblock \showarticletitle{High-Resolution Image Synthesis and Semantic
  Manipulation with Conditional {GANs}}. In \bibinfo{booktitle}{\emph{IEEE
  Conference on Computer Vision and Pattern Recognition (CVPR)}}.
\newblock


\bibitem[\protect\citeauthoryear{Wang, Mallya, and Liu}{Wang
  et~al\mbox{.}}{2021}]%
        {wang2021facevid2vid}
\bibfield{author}{\bibinfo{person}{Ting-Chun Wang}, \bibinfo{person}{Arun
  Mallya}, {and} \bibinfo{person}{Ming-Yu Liu}.}
  \bibinfo{year}{2021}\natexlab{}.
\newblock \showarticletitle{One-Shot Free-View Neural Talking-Head Synthesis
  for Video Conferencing}. In \bibinfo{booktitle}{\emph{IEEE Conference on
  Computer Vision and Pattern Recognition (CVPR)}}.
\newblock


\bibitem[\protect\citeauthoryear{Wiles, Koepke, and Zisserman}{Wiles
  et~al\mbox{.}}{2018}]%
        {Wiles18}
\bibfield{author}{\bibinfo{person}{O. Wiles}, \bibinfo{person}{A.S. Koepke},
  {and} \bibinfo{person}{A. Zisserman}.} \bibinfo{year}{2018}\natexlab{}.
\newblock \showarticletitle{X2Face: A Network for Controlling Face Generation
  by using Images, Audio, and Pose Codes}. In
  \bibinfo{booktitle}{\emph{European Conference on Computer Vision}}.
\newblock


\bibitem[\protect\citeauthoryear{Wu, Zhang, Xue, Freeman, and Tenenbaum}{Wu
  et~al\mbox{.}}{2016}]%
        {wu2016learning}
\bibfield{author}{\bibinfo{person}{Jiajun Wu}, \bibinfo{person}{Chengkai
  Zhang}, \bibinfo{person}{Tianfan Xue}, \bibinfo{person}{William~T. Freeman},
  {and} \bibinfo{person}{Joshua~B. Tenenbaum}.}
  \bibinfo{year}{2016}\natexlab{}.
\newblock \showarticletitle{Learning a Probabilistic Latent Space of Object
  Shapes via {3D} Generative-Adversarial Modeling}. In
  \bibinfo{booktitle}{\emph{Advances in Neural Information Processing Systems
  (NeurIPS)}}.
\newblock


\bibitem[\protect\citeauthoryear{Xia, Zhang, Yang, Xue, Zhou, and Yang}{Xia
  et~al\mbox{.}}{2021}]%
        {xia2021gan}
\bibfield{author}{\bibinfo{person}{Weihao Xia}, \bibinfo{person}{Yulun Zhang},
  \bibinfo{person}{Yujiu Yang}, \bibinfo{person}{Jing-Hao Xue},
  \bibinfo{person}{Bolei Zhou}, {and} \bibinfo{person}{Ming-Hsuan Yang}.}
  \bibinfo{year}{2021}\natexlab{}.
\newblock \showarticletitle{GAN inversion: A survey}.
\newblock \bibinfo{journal}{\emph{arXiv preprint arXiv:2101.05278}}
  (\bibinfo{year}{2021}).
\newblock


\bibitem[\protect\citeauthoryear{Xu, Peng, Yang, Shen, and Zhou}{Xu
  et~al\mbox{.}}{2021}]%
        {xu2021volumegan}
\bibfield{author}{\bibinfo{person}{Yinghao Xu}, \bibinfo{person}{Sida Peng},
  \bibinfo{person}{Ceyuan Yang}, \bibinfo{person}{Yujun Shen}, {and}
  \bibinfo{person}{Bolei Zhou}.} \bibinfo{year}{2021}\natexlab{}.
\newblock \showarticletitle{3D-aware Image Synthesis via Learning Structural
  and Textural Representations}.
\newblock  (\bibinfo{year}{2021}).
\newblock


\bibitem[\protect\citeauthoryear{Zakharov, Ivakhnenko, Shysheya, and
  Lempitsky}{Zakharov et~al\mbox{.}}{2020}]%
        {Zakharov20}
\bibfield{author}{\bibinfo{person}{Egor Zakharov}, \bibinfo{person}{Aleksei
  Ivakhnenko}, \bibinfo{person}{Aliaksandra Shysheya}, {and}
  \bibinfo{person}{Victor Lempitsky}.} \bibinfo{year}{2020}\natexlab{}.
\newblock \showarticletitle{Fast Bi-layer Neural Synthesis of One-Shot
  Realistic Head Avatars}. In \bibinfo{booktitle}{\emph{European Conference of
  Computer Vision (ECCV)}}.
\newblock


\bibitem[\protect\citeauthoryear{Zakharov, Shysheya, Burkov, and
  Lempitsky}{Zakharov et~al\mbox{.}}{2019}]%
        {Zakharov19}
\bibfield{author}{\bibinfo{person}{Egor Zakharov}, \bibinfo{person}{Aliaksandra
  Shysheya}, \bibinfo{person}{Egor Burkov}, {and} \bibinfo{person}{Victor
  Lempitsky}.} \bibinfo{year}{2019}\natexlab{}.
\newblock \bibinfo{booktitle}{\emph{Few-Shot Adversarial Learning of Realistic
  Neural Talking Head Models}}.
\newblock
\showeprint[arXiv]{1905.08233}


\bibitem[\protect\citeauthoryear{Zhang, Isola, Efros, Shechtman, and
  Wang}{Zhang et~al\mbox{.}}{2018}]%
        {zhang2018perceptual}
\bibfield{author}{\bibinfo{person}{Richard Zhang}, \bibinfo{person}{Phillip
  Isola}, \bibinfo{person}{Alexei~A Efros}, \bibinfo{person}{Eli Shechtman},
  {and} \bibinfo{person}{Oliver Wang}.} \bibinfo{year}{2018}\natexlab{}.
\newblock \showarticletitle{The Unreasonable Effectiveness of Deep Features as
  a Perceptual Metric}. In \bibinfo{booktitle}{\emph{IEEE Conference on
  Computer Vision and Pattern Recognition (CVPR)}}.
\newblock


\bibitem[\protect\citeauthoryear{Zhou, Xie, Ni, and Tian}{Zhou
  et~al\mbox{.}}{2021}]%
        {zhou2021cips3d}
\bibfield{author}{\bibinfo{person}{Peng Zhou}, \bibinfo{person}{Lingxi Xie},
  \bibinfo{person}{Bingbing Ni}, {and} \bibinfo{person}{Qi Tian}.}
  \bibinfo{year}{2021}\natexlab{}.
\newblock \showarticletitle{{{CIPS}}-{{3D}}: A {{3D}}-{{Aware Generator}} of
  {{GANs Based}} on {{Conditionally}}-{{Independent Pixel Synthesis}}}.
\newblock \bibinfo{journal}{\emph{arXiv preprint arXiv:2110.09788}}
  (\bibinfo{year}{2021}).
\newblock


\bibitem[\protect\citeauthoryear{Zhu, Shen, Zhao, and Zhou}{Zhu
  et~al\mbox{.}}{2020}]%
        {zhu2020domain}
\bibfield{author}{\bibinfo{person}{Jiapeng Zhu}, \bibinfo{person}{Yujun Shen},
  \bibinfo{person}{Deli Zhao}, {and} \bibinfo{person}{Bolei Zhou}.}
  \bibinfo{year}{2020}\natexlab{}.
\newblock \showarticletitle{In-domain GAN Inversion for Real Image Editing}. In
  \bibinfo{booktitle}{\emph{European Conference on Computer Vision (ECCV)}}.
\newblock


\bibitem[\protect\citeauthoryear{Zhu, Zhang, Zhang, Wu, Torralba, Tenenbaum,
  and Freeman}{Zhu et~al\mbox{.}}{2018}]%
        {zhu2018visual}
\bibfield{author}{\bibinfo{person}{Jun-Yan Zhu}, \bibinfo{person}{Zhoutong
  Zhang}, \bibinfo{person}{Chengkai Zhang}, \bibinfo{person}{Jiajun Wu},
  \bibinfo{person}{Antonio Torralba}, \bibinfo{person}{Joshua~B. Tenenbaum},
  {and} \bibinfo{person}{William~T. Freeman}.} \bibinfo{year}{2018}\natexlab{}.
\newblock \showarticletitle{Visual Object Networks: Image Generation with
  Disentangled {3D} Representations}. In \bibinfo{booktitle}{\emph{Advances in
  Neural Information Processing Systems (NeurIPS)}}.
\newblock


\bibitem[\protect\citeauthoryear{Zollh{\"o}fer, Thies, Garrido, Bradley,
  Beeler, P{\'e}rez, Stamminger, Nie{\ss}ner, and Theobalt}{Zollh{\"o}fer
  et~al\mbox{.}}{2018}]%
        {zollhofer2018state}
\bibfield{author}{\bibinfo{person}{Michael Zollh{\"o}fer},
  \bibinfo{person}{Justus Thies}, \bibinfo{person}{Pablo Garrido},
  \bibinfo{person}{Derek Bradley}, \bibinfo{person}{Thabo Beeler},
  \bibinfo{person}{Patrick P{\'e}rez}, \bibinfo{person}{Marc Stamminger},
  \bibinfo{person}{Matthias Nie{\ss}ner}, {and} \bibinfo{person}{Christian
  Theobalt}.} \bibinfo{year}{2018}\natexlab{}.
\newblock \showarticletitle{State of the Art on Monocular 3D Face
  Reconstruction, Tracking, and Applications}. In
  \bibinfo{booktitle}{\emph{Computer Graphics Forum}},
  Vol.~\bibinfo{volume}{37}.
\newblock


\end{thebibliography}

\end{document}


\title{3D GAN Inversion for Controllable Portrait Image Animation\\Supplemental Material}

\author{Connor Z. Lin}
\authornote{Both authors contributed equally.}
\author{David B. Lindell}
\authornotemark[1]
\author{Eric R. Chan}
\author{Gordon Wetzstein}
\affiliation{%
  \institution{Stanford University}
  \city{Stanford}
  \state{CA}
  \country{USA}
}
\email{connorzl@stanford.edu}
\email{lindell@stanford.edu}
\thanks{Project webpage: \url{https://www.computationalimaging.org/publications/3dganinversion}}

\renewcommand\shortauthors{Lin, C. Z. et al.}

\maketitle
\thispagestyle{empty}
\fancyfoot{}

\section{Supplementary Portrait Image Animation Results}

Here we provide additional results and details on the portrait image animation examples presented in the main paper.

\paragraph{Quantitative Results.} In Table~\ref{tab:metrics}, we further breakdown the average post distance and average expression distance metrics into separate evaluations using the pre-trained models of DECA~\cite{DECA:Siggraph2021} and Deep3DFace~\cite{deng2019accurate}. We observe the same trends across both estimators, as PIRenderer achieves more accurate expressions but performs similarly to the 3D GAN when eye expressions are fixed. 

\begin{figure*}
    \centering
    \includegraphics[width=0.9\textwidth]{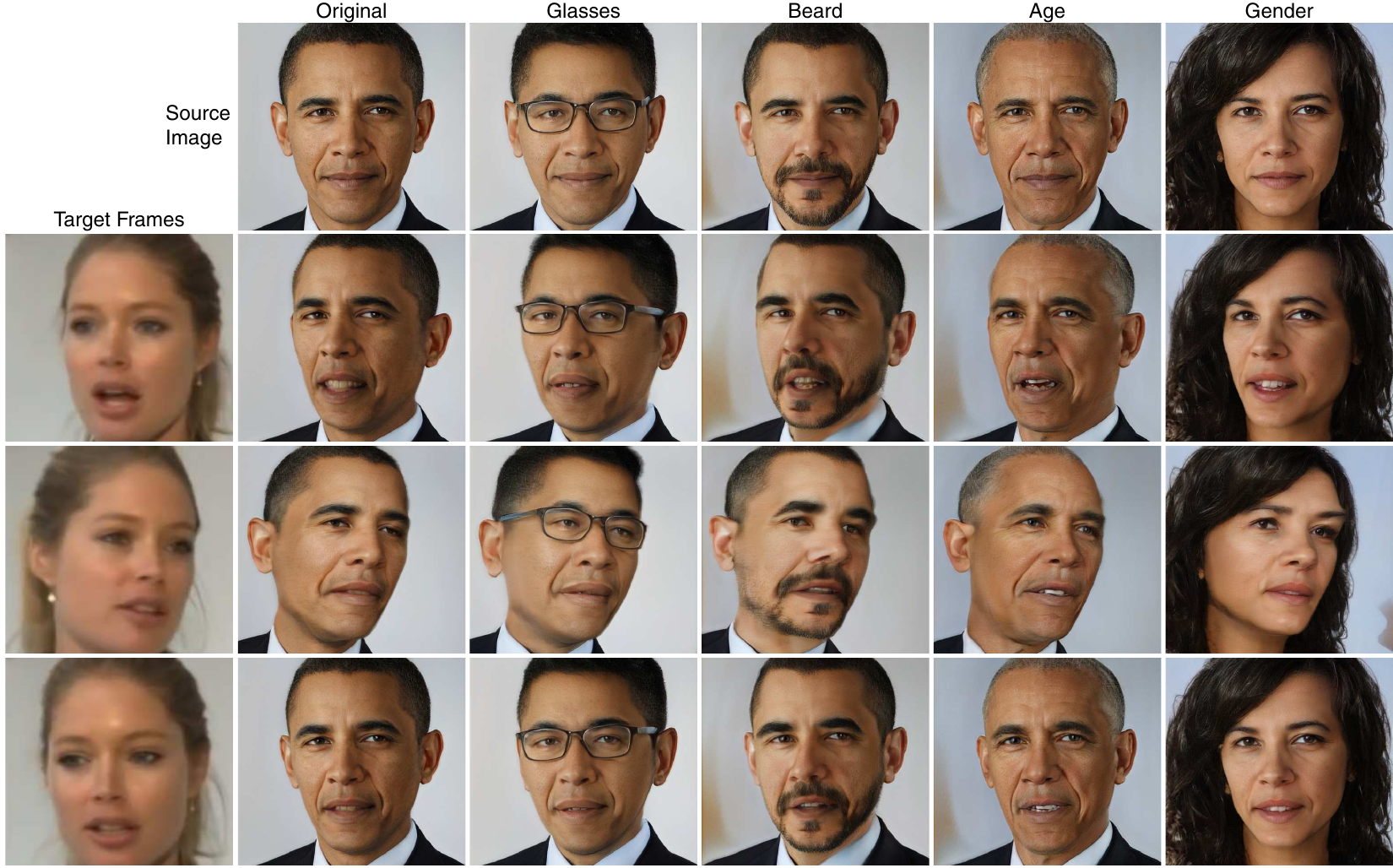}
    \caption{Portrait image animation and attribute editing using the proposed technique. Given a source portrait image and a target expression (e.g., specified with a target image), our method transfers the expression and pose to the input source image. We achieve multi-view consistent edits of pose by embedding the expression-edited portrait image into the latent space of a 3D GAN. By interpolating within the latent space of the GAN, we can also apply our method to animate attribute-edited images, allowing adjustments to age, hair, gender or appearance in addition to expression and pose.}
    \label{fig:supp_teaser}
\end{figure*}

\begin{figure*}
    \centering
    \includegraphics[width=0.9\textwidth]{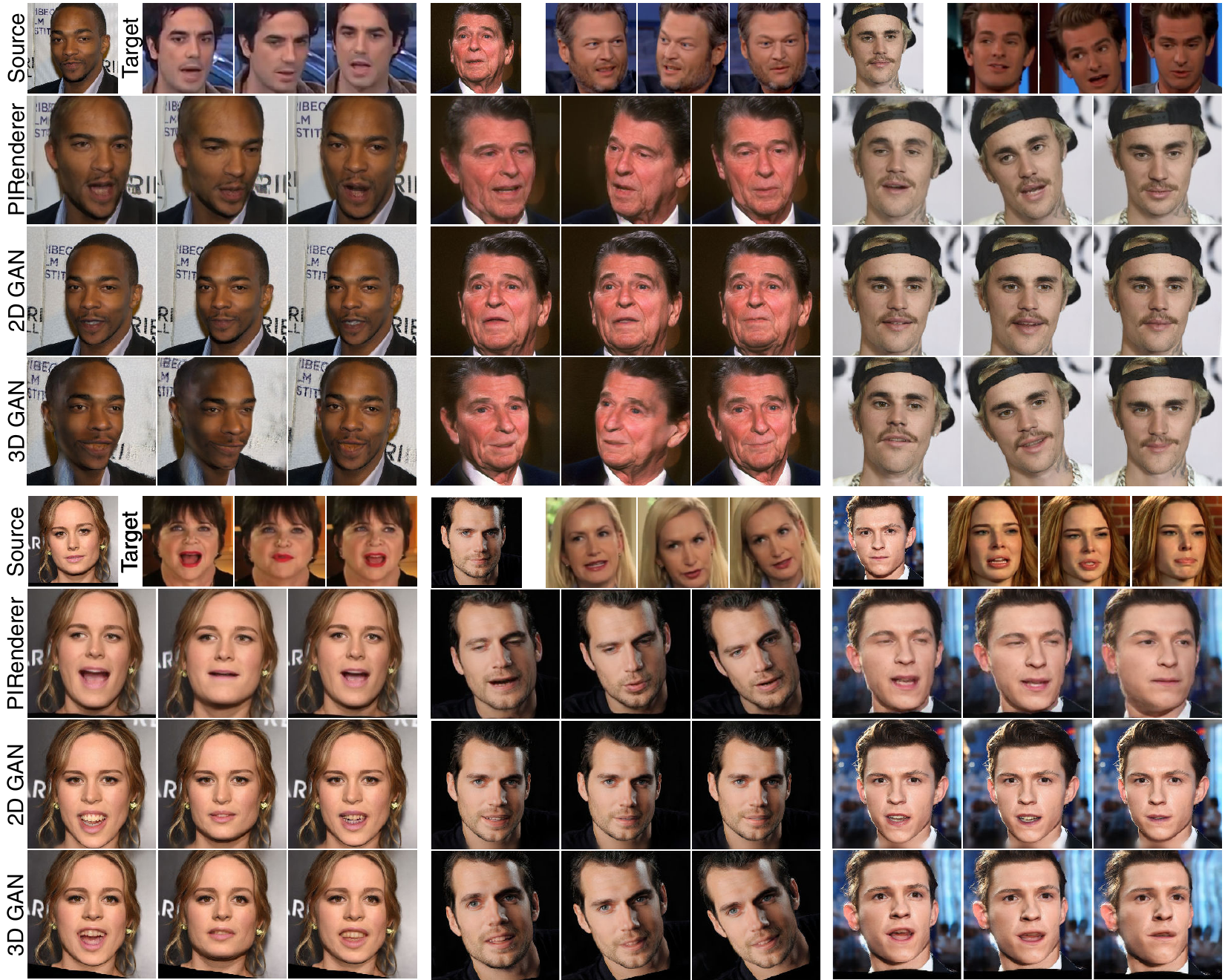}
    \caption{Portrait image animation results. We compare the proposed approach with PIRenderer~\cite{ren2021pirenderer} and our method using a pre-trained 2D GAN~\cite{karras2020stylegan2} instead of the 3D GAN. Since we cannot directly control the pose with this method, we only edit the expressions. Compared to the baselines, the proposed approach achieves better view consistency with higher output image quality compared to PIRenderer.}
    \label{fig:supp_animation}
\end{figure*}

\begin{figure*}
    \centering
    \includegraphics[width=0.9\textwidth]{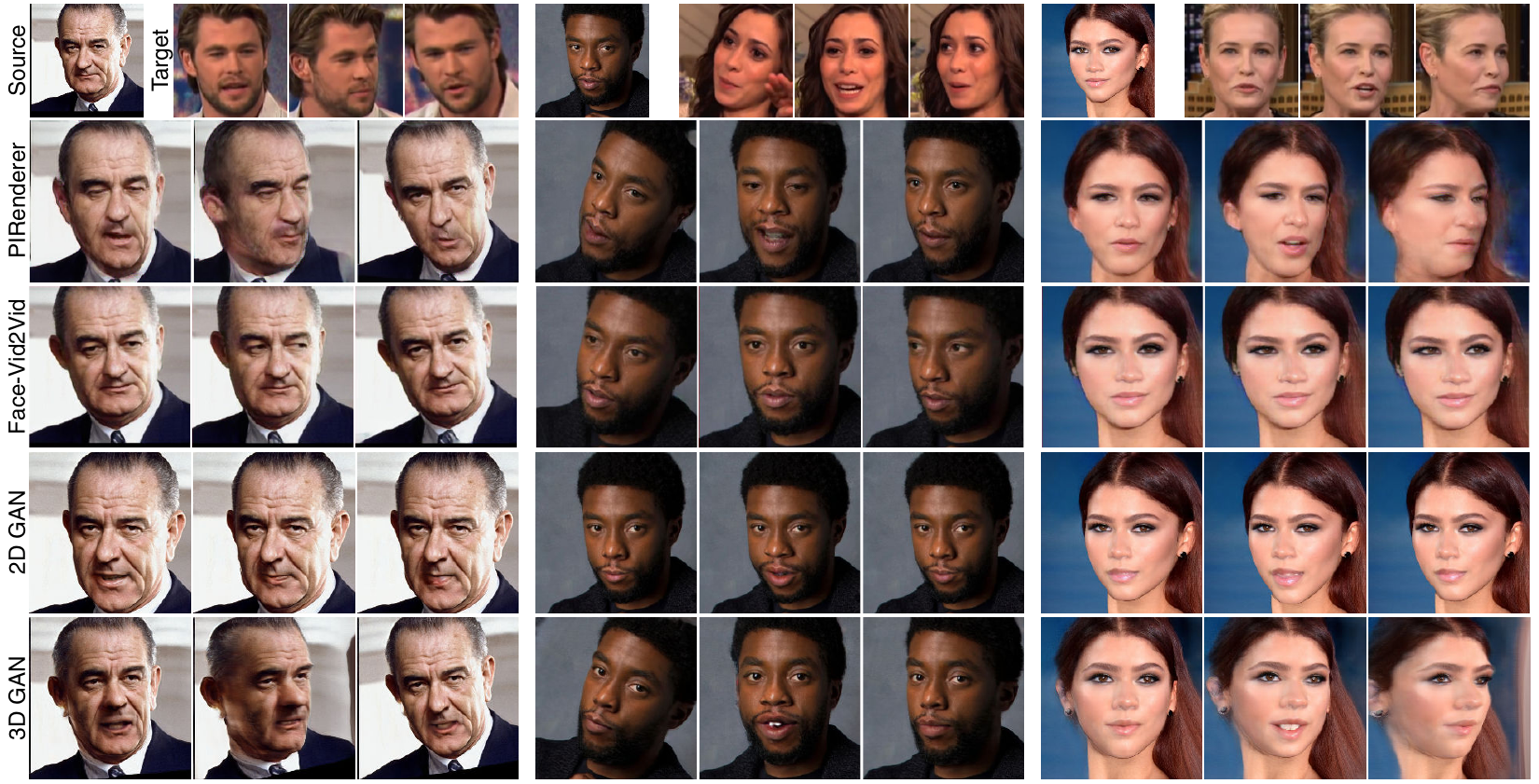}
    \caption{Portrait image animation results. We compare the proposed approach with PIRenderer~\cite{ren2021pirenderer}, Face-Vid2Vid~\cite{wang2021one}, and our method using a pre-trained 2D GAN~\cite{karras2020stylegan2} instead of the 3D GAN. Only the online demo is available for Face-Vid2Vid, so we can only control the head and eye poses. We cannot directly control the pose with the 2D GAN, so we only edit the expressions. Compared to the baselines, the proposed approach achieves better view consistency with higher output image quality compared to PIRenderer, and higher head pose and mouth expression accuracy compared to Face-Vid2Vid.}
    \label{fig:supp_vid2vid}
\end{figure*}

\paragraph{Qualitative Results.} Figure~\ref{fig:supp_teaser} demonstrates an additional result of editing an input source image and animating the result to follow the expressions of a target video. We also show more portrait image animation comparisons with PIRenderer and the 2D GAN baselines in Figure~\ref{fig:supp_animation}. Finally, Figure~\ref{fig:supp_vid2vid} shows a best-effort comparison with Face-Vid2Vid~\cite{wang2021one}, as only a limited online demo is available for controlling eye and head pose\footnote{\url{http://imaginaire.cc/vid2vid-cameo/}}. 

\section{Supplementary Attribute Editing Results}

\begin{figure*}
    \centering
    \includegraphics[width=0.9\textwidth]{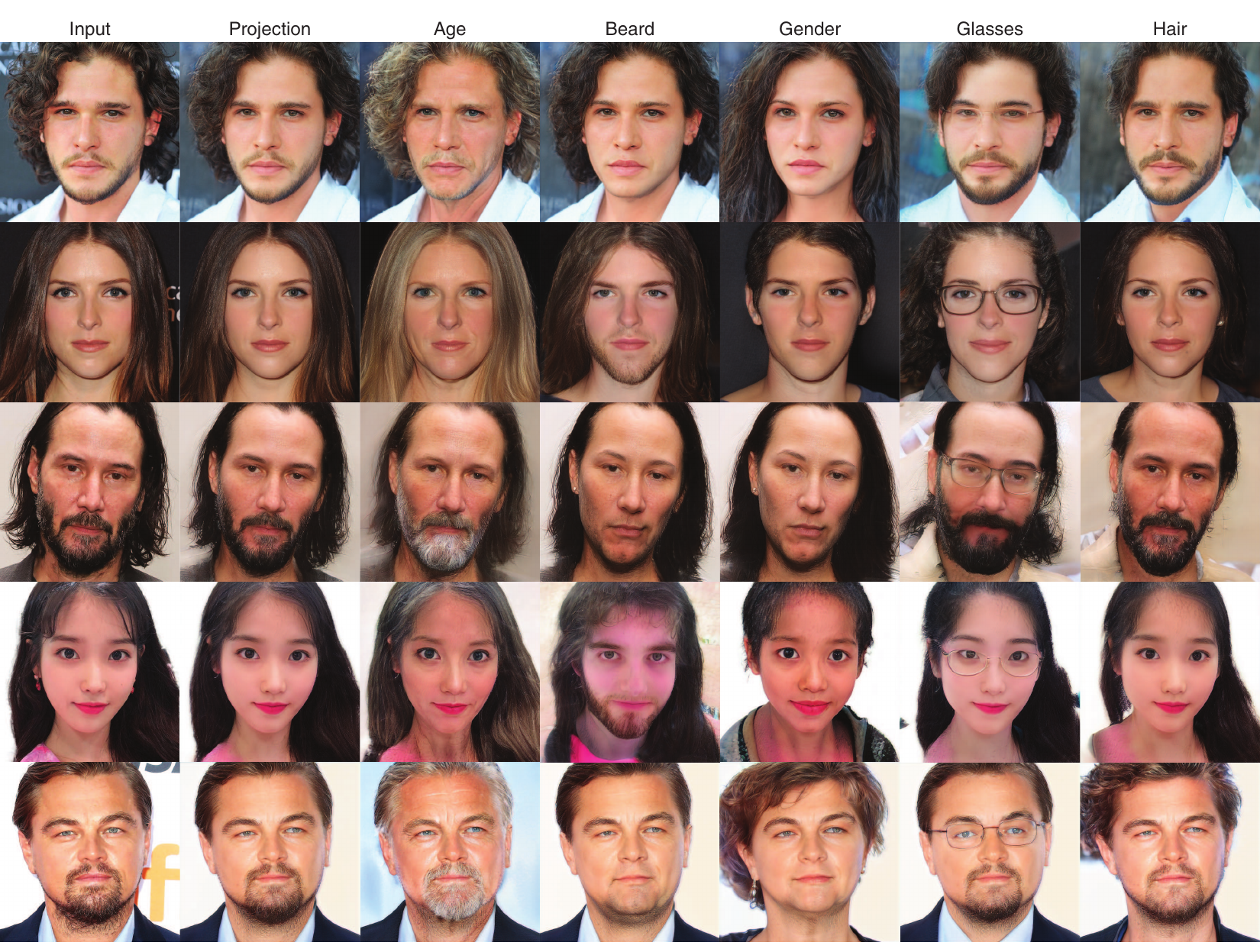}
    \caption{Portrait image attribute editing results. We train a StyleFlow model~\cite{abdal2021styleflow} using the 3D GAN latent space, then perform Pivotal Tuning Inversion~\cite{roich2021pivotal} to project real images into the latent space. Finally, we modify the conditional attribute inputs into the StyleFlow model to obtain latents capturing the desired edits. We notice that the most difficult edits are facial hair for female subjects and glasses, as both tend to incur the most identity shift -- the former involving gender and the latter involving hair.}
    \label{fig:supp_attribute}
\end{figure*}

\paragraph{Qualitative Results.} In Figure~\ref{fig:supp_attribute}, we illustrate more attribute editing results using StyleFlow~\cite{abdal2021styleflow} and the 3D GAN. Similar to the 2D GAN results shown in the original StyleFlow paper, we observe that the latent space of the 3D GAN facilitates realistic attribute editing. Facial hair edits for female subjects and glasses edits incur the most identity shift due to the sparsity of such examples in FFHQ, the 3D GAN's training dataset.

 
\begin{table*}[t]
    \centering
    \begin{tabular}{lcc|cc|cc}
    \toprule
        & \multirow{2}{*}{FID$\downarrow$}  & \multirow{2}{*}{ID $\uparrow$} & \multicolumn{2}{c|}{Avg. Pose Dist.$\downarrow$} & \multicolumn{2}{c}{Avg. Expression Dist.$\downarrow$} \\
        & & & DECA & D3DF & DECA & D3DF\\\midrule
        PIRenderer (w/o eyes, w/o pose) & 53.916         & -              & 0.183          & 0.250          & 0.361           & 0.437 \\
        PIRenderer (w/o pose)           & 53.959         & -              & 0.179          & 0.247          & 0.299           & 0.386 \\
        PIRenderer (w/o eyes)           & 63.844         & 0.694          & 0.043          & 0.039          & 0.354           & 0.424 \\
        PIRenderer                      & 64.379         & 0.700          & 0.040          & 0.040          & \textbf{0.282}  & \textbf{0.373} \\\midrule
        2D GAN (w/o pose)                         & 17.812         & -              & 0.179          & 0.246          & 0.369           & 0.434 \\\midrule
        3D GAN (w/o pose)               & \textbf{16.504}& -              & 0.180          & 0.246          & 0.369           & 0.433 \\
        3D GAN                          & 31.176         & \textbf{0.733} & \textbf{0.039} & \textbf{0.030} & 0.362           & 0.433 \\
    \bottomrule
    \end{tabular}

\caption{Quantitative evaluation of portrait image animation. We compare our approach using the pre-trained 3D GAN of Chan et al.~\shortcite{Chan2021} to our approach using a 2D GAN~\cite{karras2020stylegan2} and PIRenderer~\cite{ren2021pirenderer}. The average pose and expression distances are calculated using the estimated pose and expression encodings from the pre-trained models of DECA~\cite{DECA:Siggraph2021} and Deep3DFace~\cite{deng2019accurate}, and identity consistency (ID) is calculated using a pre-trained Arcface model~\cite{deng2018arcface}. The proposed method using the 3D GAN achieves the best accuracy in terms of FID score, average pose distance, and identity consistency. We hypothesize that the limited ability of our model to capture eye motion (e.g., blinking) results in a penalty to the expression distance metric. A similar drop in performance is observed for a variant of PIRenderer with fixed eye expressions (w/o eyes). Finally, since the 2D GAN does not model pose, we observe that its performance is similar to a variant of the 3D GAN with fixed poses or PIRenderer where poses and eye expressions are fixed. Note that methods with fixed pose (w/o pose) have improved FID scores because they are not animated away from the initial poses of the ground truth image dataset.}
    \label{tab:metrics}
\end{table*}

\clearpage 

\bibliographystyle{ACM-Reference-Format}
\bibliography{references}